%% file: main.tex
\documentclass[10pt,twocolumn,letterpaper]{article}
\usepackage[accsupp]{axessibility}  %
\usepackage{cvpr}
\usepackage{amsmath}
\usepackage{amssymb}
\usepackage{booktabs}
\usepackage{cuted}
\usepackage{capt-of}
\usepackage{graphicx}
\usepackage{xcolor}
\usepackage{comment}
\usepackage{multirow}
\usepackage{bm}
\usepackage[pagebackref,breaklinks,colorlinks]{hyperref}

\usepackage[capitalize]{cleveref}
\crefname{section}{Sec.}{Secs.}
\Crefname{section}{Section}{Sections}
\Crefname{table}{Table}{Tables}
\crefname{table}{Tab.}{Tabs.}

\newif\ifdrafting
\draftingtrue %
\ifdrafting
    
    \newcommand{\gs}[1]{{\leavevmode\color[rgb]{0,0,1}[Gengshan: #1]}}
    \newcommand{\deva}[1]{{\leavevmode\color[rgb]{1,0,1}[Deva: #1]}}
    \newcommand{\han}[1]{{\leavevmode\color[rgb]{0,0.5,0.7}[Han: #1]}}
    \newcommand{\mv}[1]{{\leavevmode\color[rgb]{0.2,0.,0.8}[Minh: #1]}}

\else
    
    \newcommand{\gs}[1]{}
    \newcommand{\han}[1]{}
    \newcommand{\mv}[1]{}
    \newcommand{\deva}[1]{}
\fi

\newcommand{\ourmethod}[1]{BANMo}

\def\cvprPaperID{3300}
\def\confName{CVPR}
\def\confYear{2022}

\title{\ourmethod{}: Building Animatable 3D Neural Models from Many Casual Videos}

\author{Gengshan Yang$^{2}$\thanks{Work done when interning at Meta AI} \; Minh Vo$^3$ \; Natalia Neverova$^{1}$ \; Deva Ramanan$^{2}$ \; Andrea Vedaldi$^{1}$ \; Hanbyul Joo$^{1}$\\[3pt]
$^1$Meta AI \quad $^2$Carnegie Mellon University \quad $^3$Meta Reality Labs \\
}

\begin{document}
\maketitle
\begin{strip}\centering
\vspace{-30pt}
\includegraphics[width=\linewidth, trim={0 0cm 0 0cm},clip]{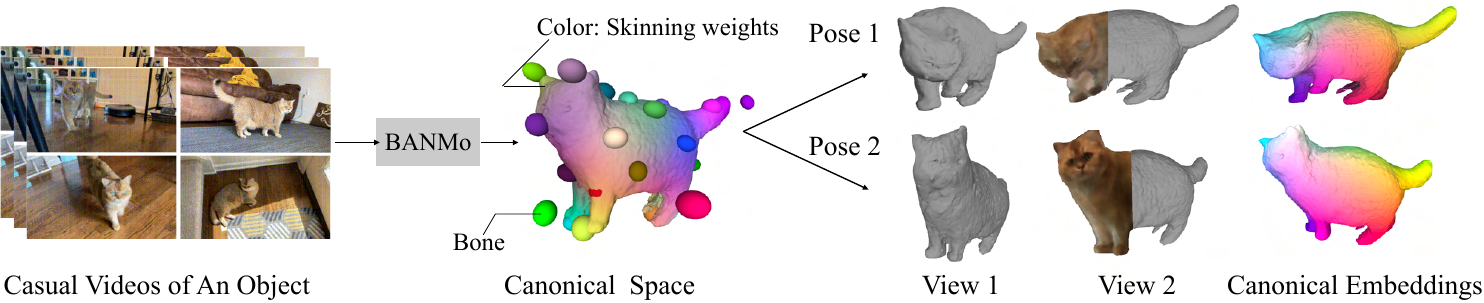}
\captionof{figure}{Given multiple casual videos capturing a deformable object, \ourmethod{} reconstructs an animatable 3D model, including an implicit canonical 3D shape, appearance, skinning weights, and time-varying articulations, without pre-defined shape templates or registered cameras. {\bf Left}: Input videos; {\bf Middle}: 3D shape, bones, and skinning weights (visualized as surface colors) in the canonical space; {\bf Right}: Posed reconstruction at each time instance with color and canonical embeddings (correspondences are shown as the same colors).
\label{fig:teaser}}
\end{strip}

\begin{abstract}
Prior work for articulated 3D shape reconstruction often relies on specialized multi-view and depth sensors or pre-built deformable 3D models. Such methods do not scale to diverse sets of objects in the wild. We present a method that requires neither of them. It aims to create high-fidelity, articulated 3D models from many casual RGB videos in a differentiable rendering framework. Our key insight is to merge three schools of thought: (1) classic deformable shape models that make use of articulated bones and blend skinning, (2) canonical embeddings that establish correspondences between pixels and a canonical 3D model, and (3) volumetric neural radiance fields (NeRFs) that are amenable to gradient-based optimization. We introduce neural blend skinning models that allow for differentiable and invertible articulated deformations. When combined with canonical embeddings, such models allow us to establish dense correspondences across videos that can be self-supervised with cycle consistency. On real and synthetic datasets, our method shows higher-fidelity 3D reconstructions than prior works for humans and animals, with the ability to render realistic images from novel viewpoints. Project page: \href{https://banmo-www.github.io/}{https://banmo-www.github.io/}.
\end{abstract}

\input{intro}

\section{Related work}
\noindent\textbf{Human and animal body models.}
A large body of work in 3D human and animal reconstruction uses parametric shape models~\cite{SMPL:2015,SMPL-X:2019,xiang2019monocular,vo2020spatiotemporal,Zuffi:CVPR:2018,Zuffi:CVPR:2017}, which are built from registered 3D scans of human or animals, and serve to recover 3D shapes given a single image or video at test time ~\cite{badger2020,biggs2020wldo,kocabas2019vibe,zuffi2019three,kocabas2019vibe}. Although parametric body models achieve great success in reconstructing human with large amounts of ground-truth 3D data, it is challenging to apply the same methodology to categories with limited 3D data, such as animals and humans in diverse sets of clothing.

\noindent\textbf{Category reconstruction from images or videos.} 
A number of recent methods build deformable 3D models of object categories from images or videos with weak 2D annotations, such as keypoints, object silhouettes, and optical flow, obtained from human annotators or predicted by off-the-shelf models~\cite{Ye_2021_CVPR,ucmrGoel20,cmrKanazawa18,li2020self, vmr2020, wu2021dove, kokkinos2021point}. Such methods often rely on a coarse shape template~\cite{kulkarni2020articulation,tulsiani2020imr,zhi2020texmesh}, and are not able to recover fine-grained details or large deformations. Recently, HDNet~\cite{Jafarian_2021_CVPR_TikTok} uses social media videos to learn depth estimators for clothed human.

\noindent\textbf{Category-agnostic video shape reconstruction.} %
Non-rigid structure from motion (NRSfM) methods~\cite{bregler2000recovering,gotardo2011non,kong2019deep,sidhu2020neural,kumar2020non} reconstruct non-rigid 3D shapes from a set of 2D point trajectories in a class-agnostic way. However, due to difficulties in obtaining accurate long-range correspondences~\cite{sand2008particle,sundaram2010dense}, they do not work well for videos in the wild. Recent efforts such as LASR and ViSER~\cite{yang2021lasr,yang2021viser} reconstruct articulated shapes from a monocular video with differentiable rendering. %
As our results show, they may still produce blurry geometry and unrealistic articulations. %

\noindent\textbf{Neural radiance fields.}
Prior works on NeRF optimize a continuous scene function for novel view synthesis given a set of images, often assuming the scene is rigid and camera poses can be accurately registered to the background~\cite{mildenhall2020nerf, martinbrualla2020nerfw,meng2021gnerf, lin2021barf, jeong2021self,wang2021nerf}. To extend NeRF to dynamic scenes, recent works introduce additional functions to deform observed points to a canonical space or over time ~\cite{pumarola2020d, park2021nerfies, li2021neural, tretschk2021nonrigid, park2021hypernerf, wang2021neural}. However, they heavily rely on background registration, and fail when the motion between objects and background is large. Moreover, the deformations cannot be explicitly controlled by user inputs. Similar to our goal, some recent works~\cite{peng2021animatable, su2021anerf, 2021narf, peng2021neural, liu2021neural} produce pose-controllable NeRFs, but they rely on a human body model, or synchronized multi-view video inputs.

\begin{figure*}[ht!]
    \centering
     \includegraphics[width=0.9\linewidth, trim={0 0 0 0},clip]{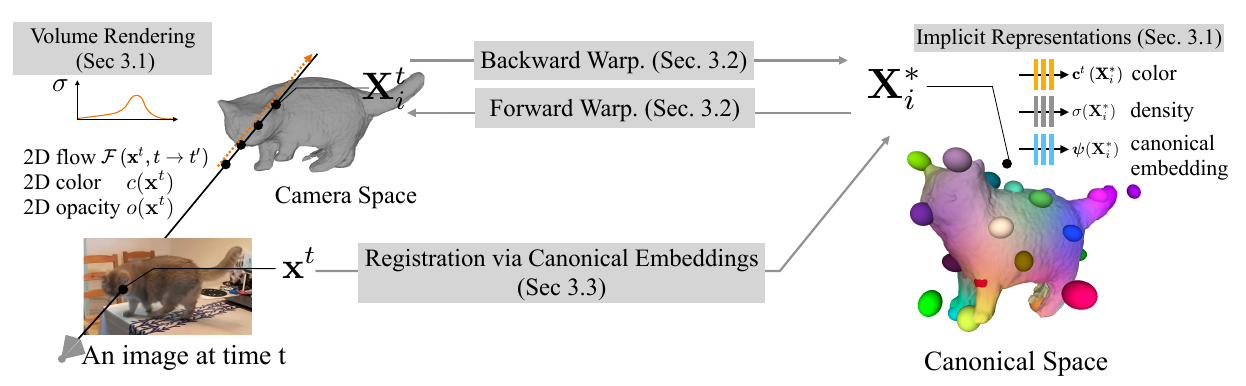}
    \caption{\textbf{Method overview.} BANMo optimizes a set of shape and deformation parameters (Sec.~\ref{sec:model}) that describe the video observations in pixel colors, silhouettes, optical flow, and higher-order features descriptors, based on a differentiable volume rendering framework. BANMo uses a neural blend skinning model (Sec.~\ref{sec:lbs}) to transform 3D points between the camera space and the canonical space, enabling handling large deformations. To register pixels across videos, BANMo jointly optimizes an implicit canonical embedding (CE) (Sec.~\ref{sec:registration}). 
    }
    \label{fig:overview}
\vspace{-10pt}
\end{figure*}

\input{method}

\section{Experiments}
\label{sec:exp}

\noindent{\bf Implementation details.}
Our implementation of implicit shape and appearance models follows NeRF~\cite{mildenhall2020nerf}, except that our shape model outputs SDF, which is transformed to density for volume rendering.  To extract the rest surface, we find the zero-level set of SDF by running marching cubes on a $256^3$ grid. To obtain articulated shapes at each time instance, we articulate points on the rest surface with forward deformation $\mathcal{W}^{t,\rightarrow}$. 

\noindent{\bf Optimization details.}
We initialize $\mathbf{MLP}_{\textrm{SDF}}$ such that it approximates a unit sphere~\cite{yariv2020multiview}. We use $B=25$ rest bones, which are initialized with unit scale, identity orientation, and centers uniformly spaced on the initial rest surface. During optimization, we reinitialize the rest bones at $\{20\%,67\%\}$ of total iterations and further encourage them to stay close to the surface with a Sinkhorn divergence loss~\cite{feydy2019interpolating}.
In a batch, we sample $N^I=512$ image pairs and sample $N^p=6144$ pixels from all image pairs for rendering. The interval between image pairs is randomly chosen $\Delta T\in\{1,2,4,8,16,32\}$. To stabilize optimization, we observe that $N_I$ needs to roughly match the number of input frames. The reconstruction quality improves with more iterations and we find 36k iterations (15 hours on a V100 GPU) already produces high-fidelity details. %
Please find a list of hyper-parameters in supplement.

\subsection{Dataset and Metrics}
\label{sec:exp1}
\noindent{\bf Qualitative: Casual videos dataset.} We demonstrate \ourmethod{}'s ability to reconstruct 3D models from casual videos of animals and humans. Object silhouette and optical flow (for computing reconstruction losses Eq.~\ref{eq:flow}) are extracted by off-the-shelf models, PointRend and VCN-robust~\cite{kirillov2020pointrend, yang2019volumetric}. Two special challenges arise from the casual nature of the video captures. First, each video collection contains around 1k images, an order of magnitudes larger those used in prior work~\cite{park2021nerfies, li2021neural, yang2021viser, mildenhall2020nerf}, which requires the method to handle reconstructions at a larger scale. Second, the dataset makes no control over camera movement or object 
movement. In particular, objects freely moves in a video and background changes across videos, posing challenges to standard SfM pipelines to estimate the object root poses. We show results on 11 videos (totaling 900 images) of a British shorthair cat denoted as \texttt{casual-cat} below. Please find other results in the project webpage.

\noindent{\bf Quantitative: AMA human dataset.} Articulated Mesh Animation (AMA) dataset~\cite{Vlasic-08} contains multi-view videos captured by 8 synchronized cameras. It provides high-fidelity ground-truth meshes with clothing. We use 2 sets of videos of the same actor (\texttt{swing} and \texttt{samba}), totaling 2600 frames, as the input to optimization. We use the ground-truth object silhouettes. Time synchronization and camera extrinsics are \emph{not} used. 

\noindent{\bf Quantitative: Animated Objects dataset.}
We download free animated 3D models from TurboSquid, including an \texttt{eagle} model and a model for human \texttt{hands}. We render them from different camera trajectories with partially overlapping motions. Each animated object is rendered as 5 videos with 150 frames per video. We provide ground-truth root poses and object silhouettes to \ourmethod{} and baselines. 

\noindent{\bf Metrics.}
We quantify the results using both Chamfer distances and F-scores. 
Chamfer distance computes the average distance between the ground-truth and the estimated surface points by finding the nearest neighbour matches, but it is sensitive to outliers. Therefore, we further report the F-score at distance thresholds $d=2\%$ of the longest edge of the axis-aligned object bounding box~\cite{tatarchenko2019single}.
To account for the unknown scale and global rigid motion, we pre-align the estimated shape to the ground-truth via Iterative Closest Point (ICP) up to a 3D similarity transformation.

\subsection{Reconstruction Results}

\begin{table}
    \caption{\textbf{Quantitative results on AMA and Animated Objects.} 3D Chamfer distance (cm, $\downarrow$) and F-score (\%, $\uparrow$) averaged over all frames. The 3D models for \texttt{eagle} and \texttt{hands} are resized such that the longest edge of the axis-aligned object bounding box is 2m. \textit{* with ground-truth root poses.} $S$: single-video results. All methods are assigned with the same initial root pose.}
    \small
    \centering
    \begin{tabular}{lcccccc}
	\toprule
	\multirow{2.5}{*}{Method}
    &\multicolumn{2}{c}{\texttt{AMA-swing}}&\multicolumn{2}{c}{\texttt{Eagle$^*$}}&\multicolumn{2}{c}{\texttt{Hands$^*$}} \\
\cmidrule(lr){2-3}\cmidrule(lr){4-5}\cmidrule(lr){6-7}&CD &F@2\%&CD &F@2\%&CD &F@2\%\\
\midrule
Ours    & {\bf 9.1} & {\bf 57.0} & {\bf 8.1} & {\bf 56.7} &  {\bf 7.5} & {\bf 49.6}\\
ViSER   & 15.7& 52.2 & 23.0 & 20.6 & 16.8 & 21.3 \\
\midrule
Ours$^S$ & 9.4 & 56.8 & 10.8 & 48.6 & 10.5 & 35.2\\
Nerfies$^S$ & 22.6 & 13.2 & 18.4 & 18.0 & 24.4 & 14.9\\
\bottomrule
\label{tab:quan-synthetic}
\end{tabular}
\vspace{-20pt}
\end{table}

We show qualitative comparison in Fig.~\ref{fig:exp1-comp} and quantitative comparison in Tab.~\ref{tab:quan-synthetic}.

\noindent{\bf Baseline setup.}
Nerfies~\cite{park2021nerfies} is designed for a single continuously captured video, assuming object root body pose can be compensated by background-SfM. In our setup, object moves and background SfM does not provide root poses for the object. When focused on the deformable object, SfM (such as COLMAP) failed to converge due to violation of rigidity, leading to very few successful registrations (18 over 900 images registered on \texttt{casual-cat}). To make a fair comparison, we provide Nerfies with rough initial root poses (obtained from our $\mathrm{PoseNet}$, Sec.~\ref{sec:opt}). After optimization, meshes are extracted by running marching cubes on a $256^3$ grid. Another baseline, ViSER~\cite{yang2021viser}, directly optimizes object shape and poses using optical flow, silhouette, and color reconstruction losses. It does not assume category-level priors such as CSE features, and therefore applicable to generic object categories. However, ViSER's root pose estimation is sensitive to large deformation and a large number of input frames (more than 20). Since it produces worse results than our $\mathrm{PoseNet}$, we provide ViSER the same root poses from our initialization pipeline.

\noindent{\bf Comparison with Nerfies.}  Nerfies optimizes SE(3) fields with photometric error, which fails at large motion and fails to register pixels across videos. In contrast, BANMo optimizes an articulated bones model using \emph{featuremetric} consistency w.r.t. a
pre-trained CSE feature embedding. As shown in Fig.~\ref{fig:exp1-comp}, although single-video Nerfies reconstructs reasonable 3D shapes of moving objects given rough initial root pose, it fails to reconstruct large articulations, such as the fast motion of the cat's head (2nd row). Furthermore, as shown in Fig.~\ref{fig:acc-vs-num}, Nerfies is not able to leverage more videos to improve the reconstruction quality, while the reconstruction of \ourmethod{} improves given more videos. The results in Tab.~\ref{tab:quan-synthetic} suggests \ourmethod{} produces more accurate geometry than Nerfies for all sequences.

\begin{figure}[t!]
    \centering
     \includegraphics[width=0.95\linewidth, trim={0cm 4cm 0cm 4cm},clip]{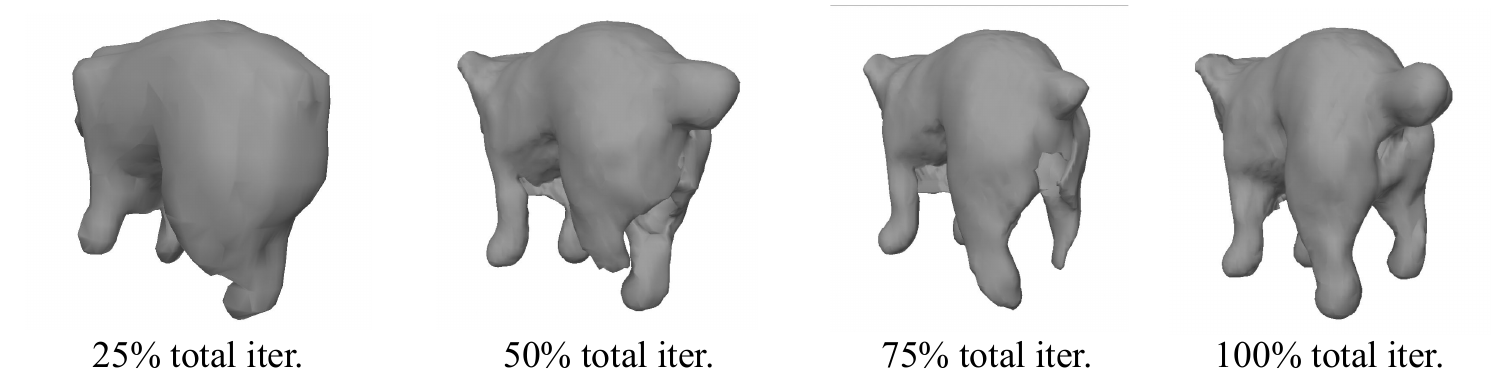}
    \caption{\textbf{Compliance to topology changes in optimization.} \ourmethod{} incorrectly reconstructs a single rear leg of the dog, but automatically corrects the topology with gradient updates.} 
    \label{fig:implicit-shape}
    \vspace{-10pt}
\end{figure}

\noindent{\bf Comparison with ViSER.}  As shown in Fig.~\ref{fig:exp1-comp}, ViSER produces reasonable articulated shapes. However, detailed geometry, such as ears, eyes, nose and rear limbs of the cat are blurred out. Furthermore, detailed articulation, such as head rotation and leg switching are not recovered. In contrast, \ourmethod{} faithfully recovers these high-fidelity geometry and motion. We observed that the neural implicit volume representation is compliant to topology changes during gradient updates (see Fig.~\ref{fig:implicit-shape}), and is therefore able to recover from bad local optima. In contrast, sub-optimal topology that happens during optimization, such as inverted faces, prevents ViSER to improve given more iterations. Compared to meshes with finite number of vertices, implicit shape representation maintains a continuous geometry, enabling us to recover detailed shape without additional cost in rendering high-res meshes.

\subsection{Diagnostics}\label{sec:aba}
We ablate the importance of each component, by using a subset of videos. To also ablate root pose initialization and registration, we test on AMA's \texttt{samba} and \texttt{swing} (325 frames in total). We include exhaustive ablations in supplement, and only highlight crucial aspects of \ourmethod{} below.

\begin{figure}[ht!]
    \centering
    \includegraphics[width=\linewidth, trim={0cm 3cm 0cm 3cm},clip]{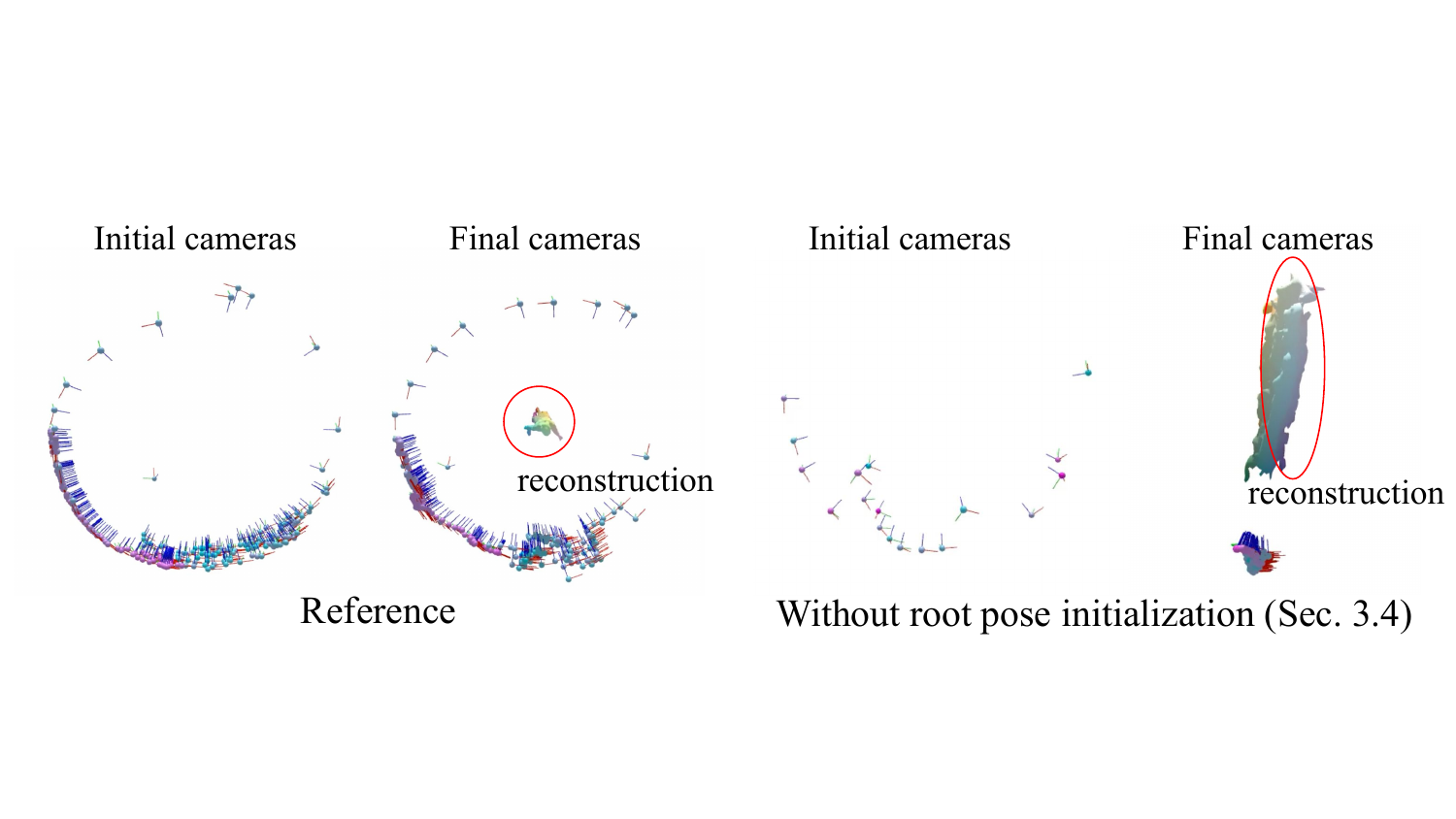}
    \caption{\textbf{Diagnostics of root pose initialization (Sec.\ref{sec:opt}).} With randomly initialized root poses, the estimated poses (on the right) collapsed to a degenerate solution, causing reconstruction to fail.}
    \label{fig:aba-init}
\end{figure}

\noindent{\bf Root pose initialization.} We show the effect of $\mathrm{PoseNet}$ for root pose initialization (Sec~\ref{sec:opt}) in Fig.~\ref{fig:aba-init}: without it, the root poses (or equivalently camera poses) collapsed to a degenerate solution after optimization.

\begin{figure}[ht!]
    \centering
     \includegraphics[width=\linewidth, trim={0cm 0cm 0cm 4.9cm},clip]{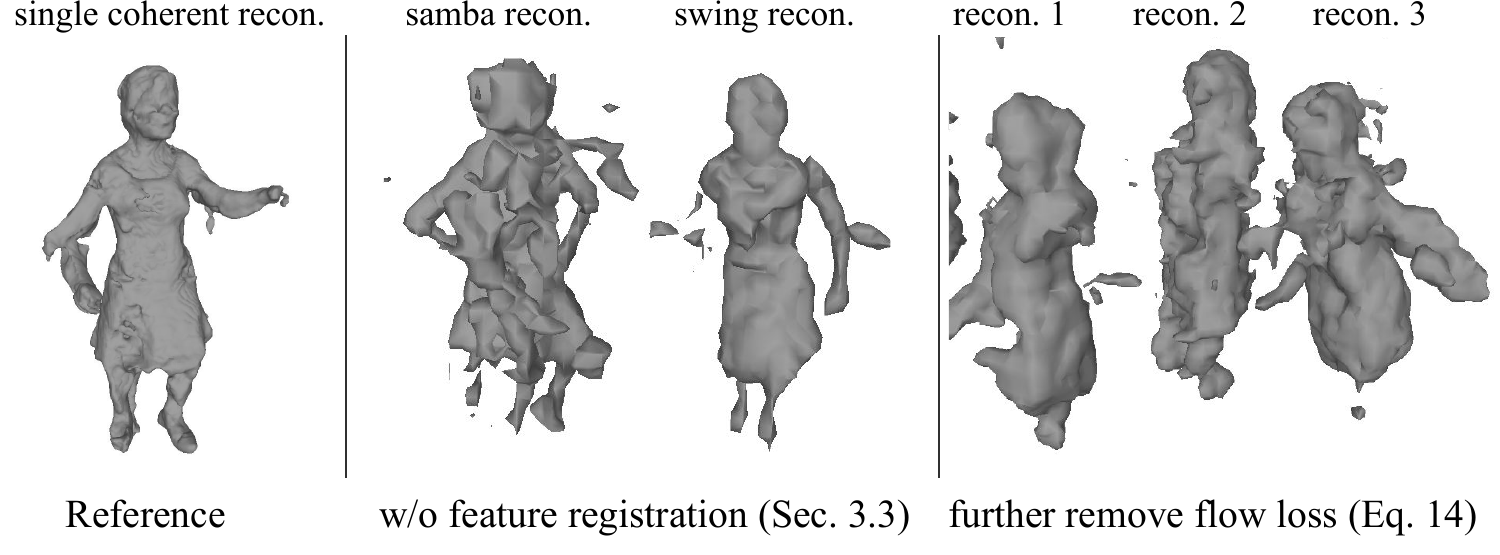}
    \caption{\textbf{Diagnostics of registration (Sec.~\ref{sec:registration}).} Without canonical embeddings (middle) or flow loss (right), our method fails to register frames to a canonical model, creating ghosting effects.}
    \label{fig:aba-registration}
\end{figure}

\noindent{\bf Registration.}  In Fig.~\ref{fig:aba-registration}, we show the benefit of using canonical embeddings (Sec~\ref{sec:registration}), and measured 2D flow (Eq.~\ref{eq:flow}) to register observations across videos and within a video. Without the canonical embeddings and corresponding losses (Eq.~\ref{eq:match}-\ref{eq:2dcyc}), each video will be reconstructed separately. With no flow reconstruction loss, multiple ghosting structures are reconstructed due to failed registration.

\noindent{\bf Deformation modeling.}
We demonstrate the benefit of using our neural blend skinning model (Sec~\ref{sec:lbs}) on an \texttt{eagle} sequence, which is challenging due to its large wing articulations. If we swap neural blend skinning for MLP-SE(3)~\cite{park2021nerfies}, the reconstruction is less regular. If we swap for MLP-translation~\cite{li2021neural,pumarola2020d}, we observe ghosting wings due to wrong geometric registration (caused by large motion). Our method can model large articulations thanks to the regularization from the Gaussian component, and also handle complex deformation such as close contact of hands.

\begin{figure}[t!]
    \centering
    \includegraphics[width=\linewidth, trim={0cm 5cm 0cm 4cm},clip]{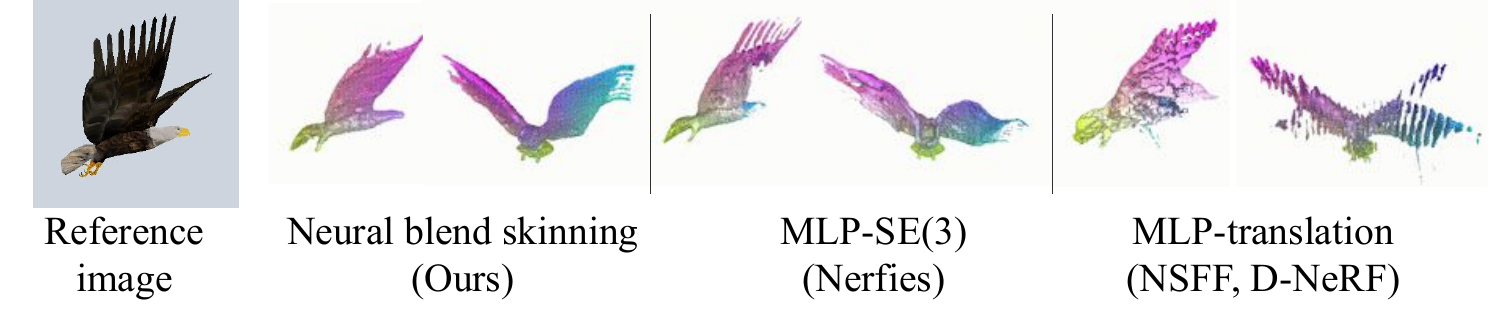}
    \caption{\textbf{Diagnostics of deformation modeling (Sec.\ref{sec:lbs}).} Replacing our neural blend skinning with MLP-SE(3) results in less regular deformation in the non-visible region. Replacing our neural blend skinning with MLP-translation as in NSFF and D-Nerf results in reconstructing ghosting wings due to significant motion.}
    \label{fig:aba-deformation}
    \vspace{-10pt}
\end{figure}

\begin{figure}[t!]
    \centering
     \includegraphics[width=0.95\linewidth, trim={0cm 0cm 0cm 0cm},clip]{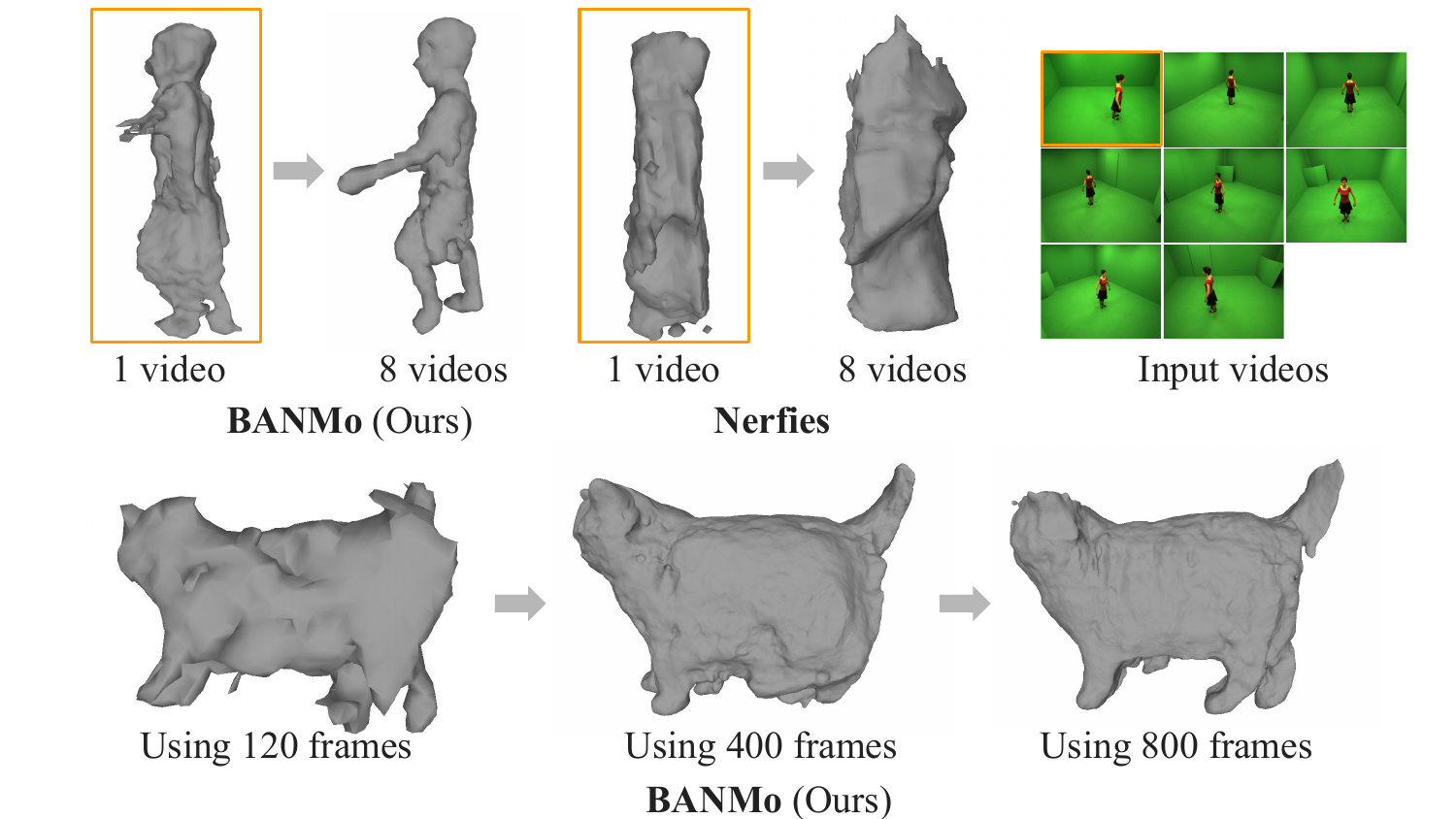}
    \caption{\textbf{Reconstruction completeness vs number of input videos and video frames.} \ourmethod{} is capable of registering more input videos if they are available, improving the reconstruction.} %
    \label{fig:acc-vs-num}
    \vspace{-12pt}
\end{figure}

\noindent{\bf Ability to leverage more videos.}
We compare \ourmethod{} to Nerfies in its ability to leverage more available video observations. To demonstrate this, we compare the reconstruction quality of optimizing 1 video vs. 8 videos from the AMA \texttt{samba} sequences. As shown in  Fig.~\ref{fig:acc-vs-num}, given more videos, our method can register them to the same canonical space, improving the reconstruction completeness and reducing shape ambiguities. In contrast, Nerfies does not produce better results given more video observations.

\begin{figure}[ht!]
    \centering
    \includegraphics[width=\linewidth, trim={0cm 4.8cm 0cm 4cm},clip]{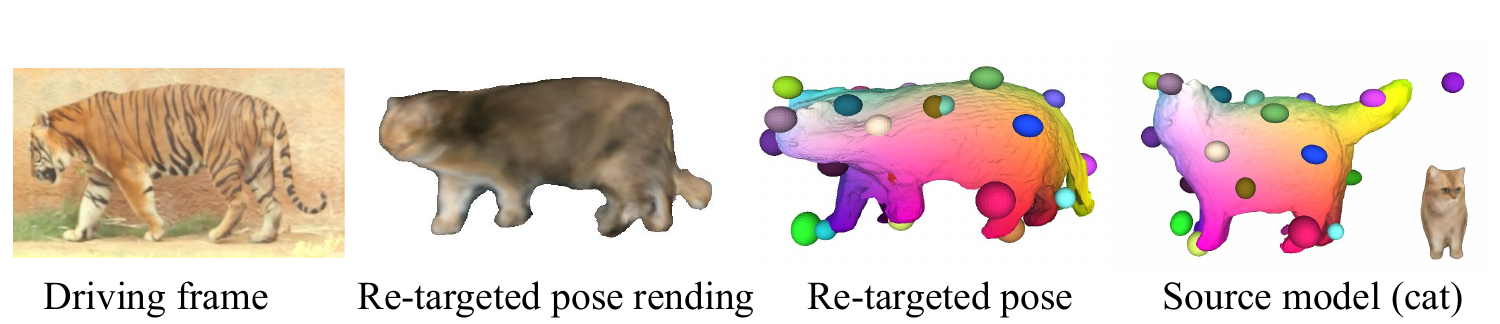}
    \caption{\textbf{Motion re-targeting from a pre-optimized cat model to a tiger.} Color coded by point locations in the canonical space.}
    \label{fig:retarget}
    \vspace{-10pt}
\end{figure}

\subsection{Application: Motion retargeting}
As a distinctive application, we demonstrate \ourmethod{}'s ability of  to retarget the articulations of a driving video to our 3D model by optimizing the frame-specific root and body pose codes $\boldsymbol{\omega}_r^t$, $\boldsymbol{\omega}_b^t$, as shown in Fig.~\ref{fig:retarget}. To do so, we first optimize all parameters over a set of \emph{training} videos from our \texttt{casual-cat} dataset. Given a driving video of a tiger, we freeze the shared model parameters (including shape, skinning, and canonical features) of the cat model, and only optimize the video-specific and frame-specific codes, i.e. root and body pose codes $\boldsymbol{\omega}_r^t$, $\boldsymbol{\omega}_b^t$, as well as the environment lighting code $\boldsymbol{\omega}_e^t$.
\vspace{-3pt}
\section{Discussion}
\vspace{-3pt}
We have presented \ourmethod{}, a method to reconstruct high-fidelity animatable 3D models from a collection of casual videos, without requiring a pre-defined shape template or pre-registered cameras. \ourmethod{} registers thousands of \emph{unsynchronized} video frames to the same canonical space by enforcing feature-metric consistencies via 2D-3D correspondence matching. We have also shown benefits of using a hybrid neural blend-skinning model for reconstructing large deformations and fine geometry. 

\noindent{\bf Limitations.} BANMo uses a pre-trained DensePose-CSE (with 2D keypoint annotaionts~\cite{neverova2021discovering}) to provide rough root body pose registration. To build generic pipelines of deformable 3D model reconstruction, a generic relative root pose estimator is needed. Similar to other works in differentiable rendering, \ourmethod{} requires a lot of compute, which increases linearly with the number of input images. We leave speeding up the optimization as future work.

\input{supp}

\clearpage
\newpage
{\small
\bibliographystyle{ieee_fullname}
\bibliography{egbib}
}

\end{document}

%% file: intro.tex
\section{Introduction}\label{sec:intro}

We are interested in developing tools that can reconstruct accurate and animatable models of 3D objects from casually collected videos.
A representative application is content creation for virtual and augmented reality, where the goal is to 3D-ify images and videos captured by users for consumption in a 3D space or creating animatable assets such as avatars.
For rigid scenes, traditional Structure from Motion (SfM) approaches can be used to leverage large collection of uncontrolled images, such as images downloaded form the web, to build accurate 3D models of landmarks and entire cities~\cite{snavely2006photo, snavely2008modeling,agarwal2011building}.
However, these approaches do not generalize to deformable objects such as family members, friends or pets, which are often the focus of user content.

We are thus interested in reconstructing 3D deformable objects from \emph{casually collected videos}.
However, individual videos may not contain sufficient information to obtain good reconstruction of a given subject.
Fortunately, we can expect that users may collect several videos of the same subjects, such as filming a family member or a pet over the span of several months or years. In this case, we wish our system to gather information from \emph{all available videos} into a single 3D model, bridging any time discontinuity.

In this paper, we present \textbf{\ourmethod{}}, a \textbf{B}uilder of \textbf{A}nimatable 3D \textbf{N}eural \textbf{Mo}dels from multiple casual RGB videos. By consolidating the 2D cues from thousands of images into a fixed canonical space, \ourmethod{} learns a high-fidelity neural implicit model for appearance, 3D shape, and articulations of the target non-rigid object. The articulation of the output model of \ourmethod{} is expressed by a neural blend skinning model, similar to \cite{yang2021lasr,yang2021viser,chen2021snarf, Saito:CVPR:2021}, making the output \emph{animatable} by manipulating bone transformations.

As shown in NRSfM~\cite{bregler2000recovering}, reconstructing a freely moving non-rigid object from monocular video is challenging, where epipolar constraints are not directly applicable. We address three core challenges in \ourmethod{}: (1) how to represent 3D geometry and appearance of the target in a canonical space; (2) how to deform 3D points between canonical space and individual time instances; (3) how to find pixel or part correspondences over videos given different viewpoint, lighting, background, and object deformations.

Concretely, we utilize neural implicit functions~\cite{mildenhall2020nerf} to represent color and 3D surface in the canonical space. This representation enables higher-fidelity 3D geometry reconstruction compared to approaches based on 3D meshes~\cite{yang2021lasr,yang2021viser}. The use of neural blending skinning in \ourmethod{} provides a way to constrain the deformation space of the target object, allowing better handling of pose variations and deformations with unknown camera parameters, compared to dynamic NeRF approaches~\cite{chen2021snarf, park2021nerfies, pumarola2020d, li2021neural}. To find correspondences, we present a module that performs dense matching between pixels and an implicit feature volume. Finally, for robust and efficient optimization over a large number of video frames, we pre-train a pose network for human and quadruped animals to provide initial camera orientations. In a nutshell, \ourmethod{} presents a way to merge the recent non-rigid object reconstruction approaches~\cite{yang2021lasr,yang2021viser} in a dynamic NeRF framework~\cite{chen2021snarf, park2021nerfies, pumarola2020d, li2021neural}, to achieve higher-fidelity non-rigid object reconstruction. We show experimentally that \ourmethod{} produces higher-fidelity 3D shape details than previous state-of-the art approaches~\cite{yang2021viser}, by taking better advantage of the large number of frames in multiple videos. 

%% file: method.tex
\section{Method}\label{sec:method}
We model the deformable object in a canonical time-invariant space, i.e. the \emph{rest} body pose space, that can be transformed to the \emph{articulated} pose in the camera space at each time instance with forward mappings, and transform back with backward mappings. We use implicit functions to represent the 3D shape, color, and dense semantic embeddings of the object. Our neural 3D model can be deformed and rendered into images at each time instance via differentiable volume rendering, and optimized to ensure consistency between the rendered images and multiple cues in the observed images, including color, silhouette, optical flow, and pixel feature embeddings. We refer readers to the overview in Fig.~\ref{fig:overview} and the list of notations in supplement.

\subsection{Shape, Appearance, and Warping Model}\label{sec:model}
We first represent shape and appearance of deformable objects in a canonical time-invariant \emph{rest pose} space.

\noindent{\bf Canonical shape model.}
In order to model the shape and appearance of an object in a canonical space, we use a method inspired by Neural Radiance Fields (NeRF)~\cite{mildenhall2020nerf}.
A 3D point $\mathbf{X}^* \in \mathbb{R}^3$ in the canonical space is associated with three properties: color $\mathbf{c} \in \mathbb{R}^3$, density $\sigma \in [0,1]$, and a canonical embedding $\boldsymbol{\psi} \in \mathbb{R}^{16}$.
These properties are predicted by the Multilayer Perceptron (MLP) networks: 
\begin{align}
    {\bf c}^t &= \mathbf{MLP}_\mathbf{c}({\bf X^*},{\bf v}^t,\boldsymbol{\omega}^t_{e}), \label{eq:color}\\
    \sigma &= \Gamma_{\beta} (\mathbf{MLP}_{\textrm{SDF}}({\bf X^*})), \label{eq:density}\\
    \boldsymbol{\psi} &= \mathbf{MLP}_{\boldsymbol\psi}({\bf X^*}) \label{eq:feature}.
\end{align}
As in NeRF, color ${\bf c}^t$ also depends on a time-varying view direction ${\bf v}^t{\in}\mathbb{R}^2$ and a learnable environment code $\boldsymbol{\omega}^t_{e}{\in}\mathbb{R}^{64}$ that captures environment illumination conditions~\cite{martinbrualla2020nerfw}.

The shape is given by $\mathbf{MLP}_{\textrm{SDF}}$, computing the Signed-Distance Function (SDF) of a point to the surface. 
To convert SDF to density $\sigma$ for volume rendering, we use the cumulative of a unimodal distributuion with zero mean and $\beta$ scale, denoted as $\Gamma_{\beta}(x)$. $\beta$ is a learnable parameter that controls the solidness of the object, approaching zero for solid objects~\cite{yariv2021volume,wang2021neus}.  Prior works~\cite{yariv2021volume, wang2021neus} have explored the cumulative of Logistic and Laplace distribution respectively and we follow VolSDF~\cite{yariv2021volume} to use the accumulative of Laplace distribution. Compared with ReLU of Softplus activations used in NeRF, it provides a principled way of extracting surface as the zero level-set of the SDF.

Finally, the $\mathbf{MLP}_{\boldsymbol{\psi}}$ network maps 3D points to a canonical feature embedding $\boldsymbol{\psi}$ that can be matched by pixels from different viewpoints and lighting conditions, enabling long-range correspondence across videos. This feature can be interpreted as a variant of Continuous Surface Embeddings (CSE)~\cite{Neverova2020cse} but defined volumetrically.

\noindent{\bf Space-time warping model.}
We consider a pair of time-dependent warping functions: \textit{forward warping function} $\mathcal{W}^{t,\rightarrow}: {\bf X}^* \rightarrow {\bf X}^t$ mapping canonical location ${\bf X}^*$ to camera space location ${\bf X}^t$ at current time and
the \textit{backward warping function} $\mathcal{W}^{t,\leftarrow}: {\bf X}^t\rightarrow {\bf X}^*$ for inverse mapping.

Prior work such as Nerfies~\cite{park2021nerfies} and Neural Scene Flow Fields (NSFF)~\cite{li2021neural} learn deformations under the assumptions of known camera poses and small object deformation.
As detailed in Sec.~\ref{sec:lbs} and Sec.~\ref{sec:opt}, 
we do not make such assumptions; instead, we adopt a neural blend-skinning model that can handle large deformations without a pre-defined skeleton model.

\noindent{\bf Volume rendering.} To render images, we use volume rendering in NeRF~\cite{mildenhall2020nerf}, but warp the 3D ray to account for deformation~\cite{park2021nerfies}.
Specifically, let $\mathbf{x}^t \in \mathbb{R}^2$ be the pixel location at time $t$, and $\mathbf{X}^t_i \in \mathbb{R}^3$ be the $i$-th 3D point sampled along the ray emanates from $\mathbf{x}^t$. As the color and density are defined in the canonical space, we \emph{pull back} the sampled points to the canonical space using
$
    {\bf X}_i^* = \mathcal{W}^{t,\leftarrow} \left( \mathbf{X}^t_i \right).
$
The color~${\bf c}$ and the opacity $o \in [0,1]$ are then given by:
\begin{equation}
    {\bf c} (\mathbf{x}^t) = \sum_{i=1}^N\tau_i {\bf c}^t_i,~~
    o (\mathbf{x}^t) = \sum_{i=1}^N\tau_i,
\end{equation}
where $N$ is the number of samples and $\tau_i$ is the free-flight probability that a photon travels between the camera center and the $i$-th sample, as given by $\tau_i=\prod_{j=1}^{i-1} p_j(1-p_i).$
Here $p_i = \exp \left(-\sigma_{i} \delta_{i}\right)$ is the probability that the photon is transmitted through the interval  $\delta_i$ between the $i$-th sample and the next. Color ${\bf c}_i$ and density ${\sigma}_i$ are computed by Eq.~\ref{eq:color}-\ref{eq:density}. We further compute the expected surface intersection:
\begin{align}
{\bf X}^* (\mathbf{x}^t) = \sum_{i=1}^N\tau_i  {\bf X}^*_i .
\label{eq:prediction_deform}
\end{align}
To render 2D flow, we \emph{push forward} the warped ray points to another time $t'$ via forward warping $\mathcal{W}^{t',\rightarrow}$to find its expected 2D re-projection:
\begin{align}
\mathbf{x}^{t'} = \sum_{i=1}^N\tau_i\Pi^{t'}\left( \mathcal{W}^{t',\rightarrow} \left({\bf X}^*_i\right)\right),
\end{align}
where $\Pi^{t'}$ is the projection matrix of a pinhole camera. We optimize video-specific $\Pi^{t'}$ given a rough initialization. With this, we compute a 2D flow rendering as:
\begin{align}
\mathcal{F}\left(\mathbf{x}^t, t \rightarrow t^{\prime}\right) = {{\bf x}^{t'}}- {\bf x}^t.
\label{eq:model_flow}
\end{align}

\subsection{Deformation Model via Neural Blend Skinning}\label{sec:lbs}
We define mappings $\mathcal{W}^{t,\rightarrow}$ and $\mathcal{W}^{t,\leftarrow}$ based on a neural blend skinning model approximating articulated body motion. Defining invertible warps for neural deformation representations is difficult~\cite{chen2021snarf}. Our formulation represents 3D warps as compositions of neural-weighted \emph{rigid-body transformations}, each of which is differentiable and invertible.

\noindent{\bf Blend skinning deformation.} Given a 3D point ${\bf X}^t$ at time $t$, we wish to find its corresponding 3D point ${\bf X}^*$ in the canonical space. Conceptually, ${\bf X}^*$ can be considered as points in the \emph{rest} pose at a fixed camera view point. Our formulation finds mappings between ${\bf X}^t$ and ${\bf X}^*$ by blending the rigid transformations of 3D coordinate of bones. Let ${\bf G}^t \in SE(3)$ be a transformation of the object root body from canonical space to time $t$, and ${\bf J}^t_{b} \in SE(3)$ be a rigid transformation that moves the $b$-th bone from its rest configuration to deformed state $t$, then we have
\begin{align}
    \label{eq:deformations}
    \mathbf{X}^t&=\mathcal{W}^{t,\rightarrow}({\bf X}^{*})=\mathbf{G}^t{\mathbf{J}^{t,\rightarrow} \mathbf{X}^{*}},\\
    \mathbf{X}^{*}&=\mathcal{W}^{t,\leftarrow}({\bf X}^t)=\mathbf{ J}^{t,\leftarrow}{({\mathbf{G}}^t)^{-1} \mathbf{X}^{t}},
\end{align}
where ${\bf J}^{t,\rightarrow}$ and ${\bf J}^{t,\leftarrow}$ are weighted averages of $B$ rigid transformations $\{ {\bf J}^t_b \}_{b \in \{1,\dots,B\}}$ that move the bones between rest configurations and time $t$ configurations. Following linear blend skinning deformation~\cite{skinningcourse:2014}, we have 
\begin{align}
    \label{eq:lbs}
\mathbf{J}^{t,\rightarrow}=\sum_{b=1}^{B} {\bf W}_{b}^{t,\rightarrow} \mathbf{J}^t_{b},\quad \mathbf{J}^{t,\leftarrow}=\sum_{b=1}^{B} {\bf W}_{b}^{t,\leftarrow} (\mathbf{J}^t_{b})^{-1},
\end{align}
where ${\bf W}_{b}^{t,\rightarrow}$ and ${\bf W}_{b}^{t,\leftarrow}$ represent pose-dependent skinning weights that assigns point ${\bf X}^*$ and ${\bf X}^t$ to the $b$-th bone. 

\noindent{\bf Pose representation.} 
We represent poses with angle-axis rotations and 3D translations, regressed from MLPs: 
\begin{align}
\label{eq:pose-mlp}
{\bf G}^t = \mathbf{MLP}_{\bf G}(\boldsymbol\omega_r^t),\quad
{\bf J}^t_b = \mathbf{MLP}_{\bf J}(\boldsymbol\omega_b^t)
\end{align}
where $\boldsymbol\omega_r^t$ and $\boldsymbol\omega_b^t$ are 128-dimensional latent codes for root pose and body pose at frame $t$ respectively. Compared with directly optimizing SE(3) poses, we find such over-parameterized representations converges better with stochastic first-order gradient methods. 
Instead of treating pose codes as independent parameters learned per-frame, we represent each dimension of the latent code as linear combinations of sinusoidal basis functions: 
\begin{equation}
    \omega_t^b = {\bf A}_i\mathcal{F}(t)
\end{equation}
where $\mathcal{F}(\cdot)$ is a 1D basis of sines and cosines with linearly-increasing frequencies at log-scale~\cite{tancik2020fourfeat}, and we learn separate weight matrices ${\bf A}_{i\in\{1\dots,M\}}$ for each video. The frame index $t$ is normalized by the maximum video length $\max_{i=1}^M|T_i|$. Using the temporal Fourier basis stabilizes the optimization and produces more smooth deformations.

\noindent{\bf Skinning weights.} 
Similar to SCANimate~\cite{Saito:CVPR:2021}, we define a skinning weight function $\mathcal{S}: ({\bf X},\boldsymbol{\omega}_b)\rightarrow {\bf W}\in\mathbb{R}^{B}$ that assigns ${\bf X}$ to bones given body pose code $\omega_b$. During backward mapping, we apply $\mathcal{S}$ to time $t$ points and pose codes $\boldsymbol\omega^t_b$ to compute backward skinning weights ${\bf W}^{t,\leftarrow}$.  During forward mapping, we apply the same $\mathcal{S}$ to the canonical space points and rest pose code $\boldsymbol\omega^*_b$ to compute the forward skinning weights ${\bf W}^{t,\rightarrow}$. 

Directly representing $\mathcal{S}$ as neural networks can be difficult to optimize. Therefore, we condition neural skinning weights on explicit 3D Gaussian ellipsoids that move along with the bone coordinates. Similar to LASR~\cite{yang2021lasr}, the Gaussian skinning weights are determined by the Mahalanobis distance between ${\bf X}$ and the Gaussian ellipsoids:
\begin{equation}
    {\bf W}_{\sigma} = ({\bf X}-{\bf C}_b)^T{\bf Q}_b({\bf X}-{\bf C}_b),
\end{equation}
where ${\bf C}_b\in\mathbb{R}^{B\times 3}$ are bone centers and ${\bf Q}_b = {\bf V}_b^T\boldsymbol{\Lambda}_b^0{\bf V}_b$ are the precision matrices composed by bone orientations ${\bf V}_b\in\mathbb{R}^{B\times 3\times3}$ and diagonal scale matrices $\boldsymbol{\Lambda}_b^0\in\mathbb{R}^{B\times 3\times3}$. When computing backward skinning weights, bone centers and orientations are transformed as $\begin{pmatrix}{\bf V}_b| {\bf C}_b\end{pmatrix} ={\bf J}_b\begin{pmatrix}{\bf V}^0_b | {\bf C}^0_b\end{pmatrix}$, where ${\bf J}_b$ are bone transforms in Eq.~\ref{eq:pose-mlp}. $\boldsymbol{\Lambda}_b^0$, ${\bf V}^0_b$ and ${\bf C}^0_b$ are learnable rest bone configurations.

To model the skinning weights for fine geometry, we find it helpful to add delta skinning weights after the coarse component is well-optimized. Delta skinning weights are represented as a coordinated-MLP, ${\bf W}_{\Delta}= \mathbf{MLP}_{\Delta}({\bf X},\boldsymbol\omega_b)\in \mathbb{R}^B$. The final skinning function is the sum of the coarse and fine components, normalized by a softmax function,
\begin{equation}
    {\bf W} = \mathcal{S}({\bf X}, \boldsymbol{}{\omega}_b) = \sigma_{\mathrm{softmax}}\big({\bf W}_{\sigma} +{\bf W}_{\Delta}\big).
\end{equation}
The Gaussian component regularizes the skinning weights to be spatially smooth and temporally consistent, and handles large deformations better than purely implicitly-defined ones. Furthermore, our formulation of the skinning weights are dependent on only pose status by construction, and therefore regularizes the space of skinning weights. 

\subsection{Registration via Canonical Embeddings}\label{sec:registration}

To register pixel observations at different time instances, \ourmethod{} maintains a canonical feature embedding that encodes semantic information of 3D points in the canonical space, which can be uniquely matched by the pixel features, and provide strong cues for registration via a joint optimization of shape, articulations, and embeddings. 

\begin{figure}[t!]
    \centering
    \includegraphics[width=\linewidth, trim={0cm 0.0cm 0cm 0.2cm},clip]{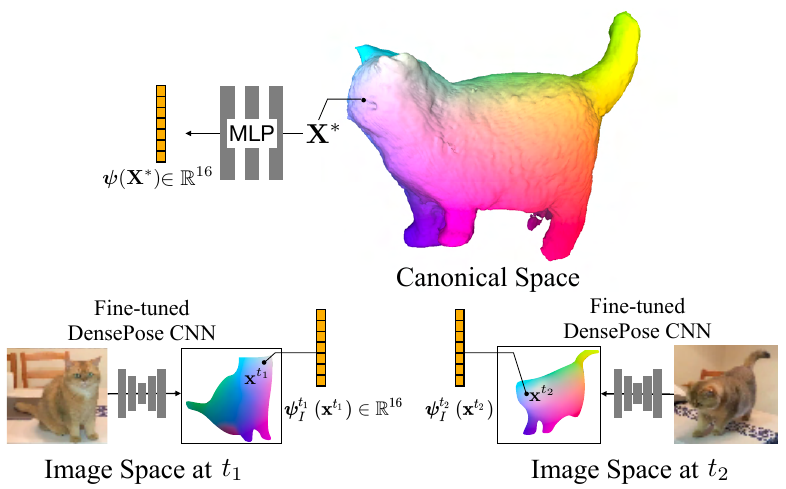}
    \caption{\textbf{Canonical Embeddings.} We jointly optimize an implicit function to produce canonical embeddings from 3D canonical points that match to the 2D DensePose CSE embeddings~\cite{Neverova2020cse}.}
    \label{fig:registration}
    \vspace{-10pt}
\end{figure}

\noindent{\bf Canonical embeddings matching.}
Given a pixel at $\mathbf{x}^t$ of frame $t$, our goal is to find a point $\mathbf{X}^*$ in the canonical space whose feature embedding  $\boldsymbol{\psi}(\mathbf{X}^*)\in \mathbb{R}^{16}$ best matches the pixel feature embedding $\boldsymbol{\psi}_I^t(\mathbf{x}^t) \in \mathbb{R}^{16}$.  The pixel embeddings $\boldsymbol{\psi}_I^t$ (of frame t) are computed by a CNN.  Different from ViSER~\cite{yang2021viser} that learns embeddings from scratch, we initialize pixel embeddings with CSE~\cite{Neverova2020cse,neverova2021discovering} that produces consistent features for semantically corresponding pixels, and optimize pixel and canonical embeddings jointly. Recall that the embedding of a canonical 3D point is computed as $\boldsymbol{\psi}(\mathbf{X}^*) =  \mathbf{MLP}_{\boldsymbol{\psi}}(\mathbf{X}^*)$ in Eq.~\ref{eq:feature}. Intuitively, $\mathbf{MLP}_{\boldsymbol{\psi}}$ is optimized to ensure the output 3D descriptor matches 2D descriptors of corresponding pixels across multiple views. To compute the 3D surface point corresponding to a 2D point $\mathbf{x}^t$, we apply soft argmax descriptor matching~\cite{luvizon2019human,kendall2017end}:
\begin{equation}
    {\hat{\bf X}^*}{(\mathbf{x}^t)} = \sum_{\mathbf{X} \in \mathbf{V^*}}\tilde{\mathbf{s}}^t(\mathbf{x}^t){\bf X},
    \label{eq:prediction_feature_matching}
\end{equation}
where $\mathbf{V}^*$ are sampled points in a tightly-bounded canonical 3D grid, and $\tilde{\bf s}$ is a normalized distribution of feature matches over the 3D grid:
$\tilde{\bf s}^t(\mathbf{x}^t) = \sigma_{\mathrm{softmax}}\Big(\alpha_s  \big\langle \boldsymbol{\psi}_I^t(\mathbf{x}^t),\boldsymbol{\psi}(\mathbf{X})\big\rangle\Big)$, where $\alpha_s$ is a learnable scaling to control the peakness of the softmax function and $\big\langle .,.\big\rangle$ is the cosine similarity.

\noindent{\bf Self-supervised canonical embedding learning.} As describe later in Eq.~\ref{eq:match}-\ref{eq:2dcyc}, the canonical embedding is self-supervised by enforcing the consistency between feature matching and geometric warping. By jointly optimizing the shape and articulation parameters via consistency losses, canonical embeddings provide strong cues to register pixels from different time instance to the canonical 3D space, and enforce a coherent reconstruction given observations from multiple videos, as validated in ablation studies (Sec.~\ref{sec:aba}).

\subsection{Optimization}\label{sec:opt}

Given multiple videos, we optimize all parameters described above, including MLPs, $\{\mathbf{MLP}_{\mathbf{c}}$,  $\mathbf{MLP}_{\textrm{SDF}}$,  $\mathbf{MLP}_{\boldsymbol{\psi}}$, $\mathbf{MLP}_{\bf G}$, $\mathbf{MLP}_{\bf J}$, $\mathbf{MLP}_{\Delta}\}$, learnable codes $\{ \boldsymbol\omega^t_{e}, \boldsymbol\omega^t_{r}, \boldsymbol\omega^t_{b}, \boldsymbol{\omega}^*_b \}$ and pixel embeddings $\boldsymbol{\psi}_I$.

\noindent{\bf Losses.}
The model is learned by minimizing three types of losses: reconstruction losses, geometric feature registration losses, and a 3D cycle-consistency regularization loss:
\begin{equation*}
    \mathcal{L} = {\underbrace {\Big(\mathcal{L}_{\textrm{sil}} + \mathcal{L}_{\textrm{rgb}} + \mathcal{L}_{\textrm{OF}} \Big)}_\text{reconstruction losses}} + {\underbrace {\Big(\mathcal{L}_{\textrm{match}} + \mathcal{L}_{\textrm{2D{\text -}cyc}} \Big)}_\text{feature registration losses}} +  \mathcal{L}_{\textrm{3D{\text -}cyc}}.
\end{equation*}
Reconstruction losses are similar to those in existing differentiable rendering pipelines~\cite{yariv2020multiview, mildenhall2020nerf}. Besides color loss $\mathcal{L}_{\textrm{rgb}}$ and silhouette loss $\mathcal{L}_{\textrm{sil}}$, we further compute flow reconstruction losses $\mathcal{L}_{\textrm{OF}}$ by comparing the rendered $\mathcal{F}$ defined in Eq.~\ref{eq:model_flow} with the observed 2D optical flow $\hat{\mathcal{F}}$ computed by an off-the-shelf flow network:
\begin{equation*}
    \mathcal{L}_{\textrm{rgb}} \!= \!\!\sum_{{\bf x}^t}\left\|{\bf c}(\mathbf{x}^t) -\hat{{\bf c}}({\bf x}^t)\right\|^2\!,\;
\mathcal{L}_{\textrm{sil}}\! =\!\!\sum_{{\bf x}^t} \left\|{\bf o}(\mathbf{x}^t) -\hat{{\bf s}}({\bf x}^t)\right\|^2\!,
\end{equation*}
\begin{equation}
\mathcal{L}_{\textrm{OF}}=\!\!\sum_{{\bf x}^t, (t, t^{\prime})}\left\|\mathcal{F}\left(\mathbf{x}^t, t \rightarrow t^{\prime}\right)-\hat{\mathcal{F}}\left(\mathbf{x}^t, t \rightarrow t^{\prime}\right)\right\|^2,
\label{eq:flow}
\end{equation}
where $\hat{{\bf c}}$ and $\hat{{\bf s}}$ are observed color and silhouette. Additionally, we define feature matching losses to enforce 3D points predicted via canonical embedding $\hat{\mathbf{X}}^{*}({\bf x}^t)$ (Eq.~\ref{eq:prediction_feature_matching}) to match the prediction from backward warping (Eq.~\ref{eq:prediction_deform}):
\begin{align}
    \mathcal{L}_{\textrm{match}} =\sum_{{\bf x}^t}\left\|\hat{\mathbf{X}}^*({\bf x}^t) - \mathbf{X}^{*}({\bf x}^t)\right\|_{2}^{2}, \label{eq:match}
\end{align}
and a geometric cycle consistency loss~\cite{kulkarni2020articulation,yang2021viser} that forces the image projection after forward warping of $\hat{\mathbf{X}}^{*}({\bf x}^t)$ to land back on its original 2D coordinates:
\begin{align}
\mathcal{L}_{\textrm{2D{\text -}cyc}} = \sum_{{\bf x}^t}\left\| \Pi^t \left( \mathcal{W}^{t,\rightarrow}( \hat{\bf X}^*({\bf x}^t)) \right) - {\bf x}^t\right\|_2^2.
\label{eq:2dcyc}
\end{align}
Similar to NSFF~\cite{li2021neural}, we regularize the deformation function $\mathcal{W}^{t,\rightarrow}(\cdot)$ and $\mathcal{W}^{t,\leftarrow}(\cdot)$ by a 3D cycle consistency loss, which encourages a sampled 3D point in the camera coordinates to be backward deformed to the canonical space and forward deformed to its original location:
\begin{equation}
\mathcal{L}_{\textrm{3D{\text -}cyc}} = \sum_{i} \tau_i\left\|\mathcal{W}^{t,\rightarrow}\Big(\mathcal{W}^{t,\leftarrow}({\bf X}^t_{i})\Big) -{\bf X}^t_{i}\right\|_{2}^{2}\\,
\end{equation}
where $\tau_i$ is the opacity that weighs the sampled points so that a point near the surface receives heavier regularization.

Our optimization is highly non-linear with local minima. To improve the robustness of optimization, we consider the following initialization strategy for root body poses.

\begin{figure*}[ht!]
    \centering
    \includegraphics[width=0.9\linewidth, trim={0cm 2cm 0cm 2cm},clip]{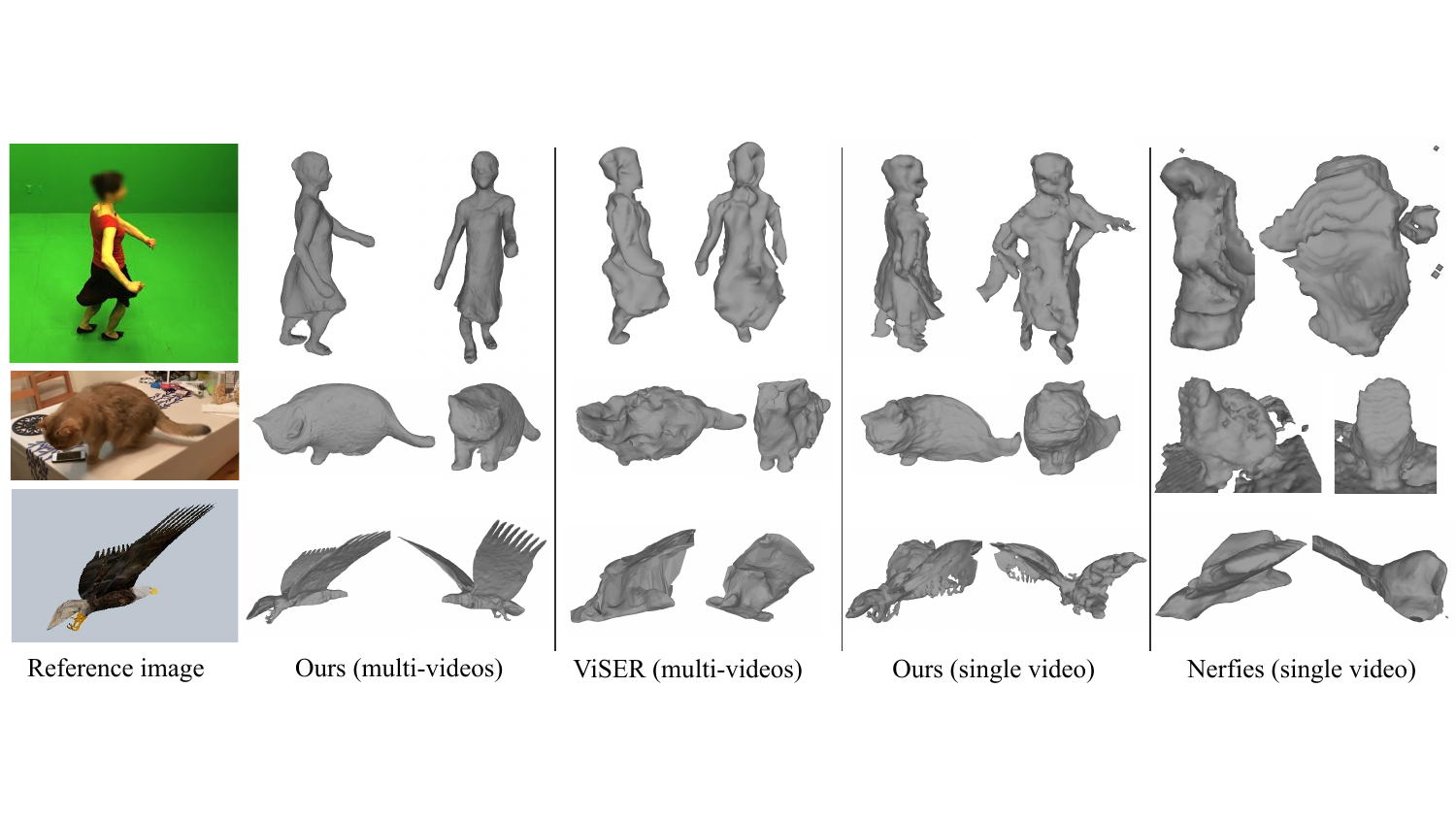}
    \caption{\textbf{Qualitative comparison of our method with prior art~\cite{yang2021viser, park2021nerfies}.} From top to bottom: AMA's \texttt{samba}, \texttt{casual-cat}, \texttt{eagle}.}
    \label{fig:exp1-comp}
\end{figure*}

\noindent{\bf Root pose initialization.}
We provide a rough per-frame initialization of root poses (${\bf G}^t$ in Eq.~\ref{eq:deformations}), similar to NeRS~\cite{zhang2021ners}. Specifically, we train a separate network $\mathrm{PoseNet}$ , which is applied to every test video frame. Similar to DenseRaC\cite{xu2019denserac}, $\mathrm{PoseNet}$ takes a DensePose CSE~\cite{Neverova2020cse} feature image as input and predicts the root pose ${\bf G}_{0}^t = \mathrm{PoseNet}(\boldsymbol{
\psi}^t_I)$, where $\boldsymbol{
\psi}^t_I\in\mathbb{R}^{112\times 112\times 16}$ is the embedding output of DensePose CSE~\cite{Neverova2020cse} from an RGB image $I_t$. We train $\mathrm{PoseNet}$ by a synthetic dataset produced offline. See supplement for details on training. Given the pre-computed ${\bf G}_{0}^t$, BANMo only needs to compute a delta root pose via MLP: 
\begin{equation}
   {\bf G}^t= \mathbf{MLP}_{\bf G}({\boldsymbol\omega^t_r}) {\bf G}_0^{t}.
\end{equation}

%% file: supp.tex
\appendix
\section{Notations}
We refer readers to a list of notations in Tab.~\ref{tab:notations} and a list of learnable parameters in Tab.~\ref{tab:parameters}. We compare with Nerfies and ViSER and summarize the differences in Tab.~\ref{tab:difference}. 

\begin{table}[h!]
    \caption{\textbf{Difference between Nerfies, ViSER, and \ourmethod{}.}}
    \vspace{-10pt}
    \small
    \centering
    \begin{tabular}{lccc}
	\toprule
Method & shape & deformation & registration\\
\midrule
Nerfies& implicit & dense SE(3) & photometric\\
ViSER  &mesh & control points & self-supervised feature\\
\ourmethod{}   &implicit & control points & pre-trained feature\\
\bottomrule
\label{tab:difference}
\end{tabular}
\vspace{-20pt}
\end{table}

\section{Method details}
\subsection{Root Pose Initialization}
\begin{figure}[ht!]
    \centering
    \includegraphics[width=\linewidth, trim={0cm 4cm 0cm 4cm},clip]{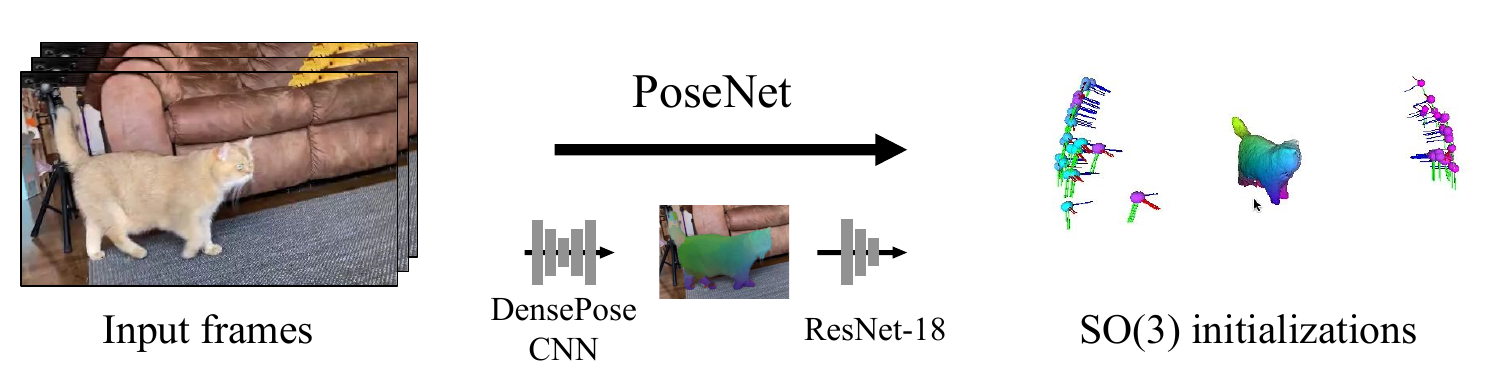}
    \caption{\textbf{Inference pipeline of PoseNet.} To initialize the optimization, we train a CNN $\mathrm{PoseNet}$ to predict root poses given a single image. $\mathrm{PoseNet}$ uses a DensePose-CNN to extract pixel features and decodes the pixel features into root pose predictions with a ResNet-18. We visualize the initial root poses on the right.  Cyan color represents earlier time stamps and magenta color represent later timestamps. }
    \label{fig:posenet}
\end{figure}

As discussed in Sec.~3.4, to make optimization robust, we train a image CNN (denoted as $\mathrm{PoseNet}$) to initialize root body transforms ${\bf G}^t$ that aligns the camera space of time $t$ to the canonical space of CSE, as shown in Fig.~
\ref{fig:posenet}.

\noindent{\bf Preliminary}
DensePose CSE~\cite{Neverova2020cse, neverova2021discovering} trains pixel embeddings $\boldsymbol{\psi}_I$ and surface feature embeddings $\boldsymbol{\psi}$ for humans and quadruped animals using 2D keypoing annotations. It represents surface embeddings by a canonical surface with $N$ vertices and vertex features $\boldsymbol{\psi}\in \mathbb{R}^{N\times16}$. A SMPL mesh is used for humans, and a sheep mesh is used for quadraped animals. The embeddings are trained such that given an pixel feature, a 3D point on the canonical surface can be uniquely located via feature matching. 

\noindent{\bf Naive PnP solution}
Given 2D-3D correspondences provided by CSE, one way to solve for ${\bf G}^t$ is to use perspective-n-points (PnP) algorithm assuming objects are rigid. However, the PnP solution suffers from catastrophic failures due to the non-rigidity of the object, which motivates our $\mathrm{PoseNet}$ solution. By training a feed-forward network with data augmentations, our  $\mathrm{PoseNet}$ solution produces fewer gross errors than the naive PnP solution.

\begin{figure}[ht!]
    \centering
    \includegraphics[width=\linewidth, trim={0cm 2cm 0cm 2cm},clip]{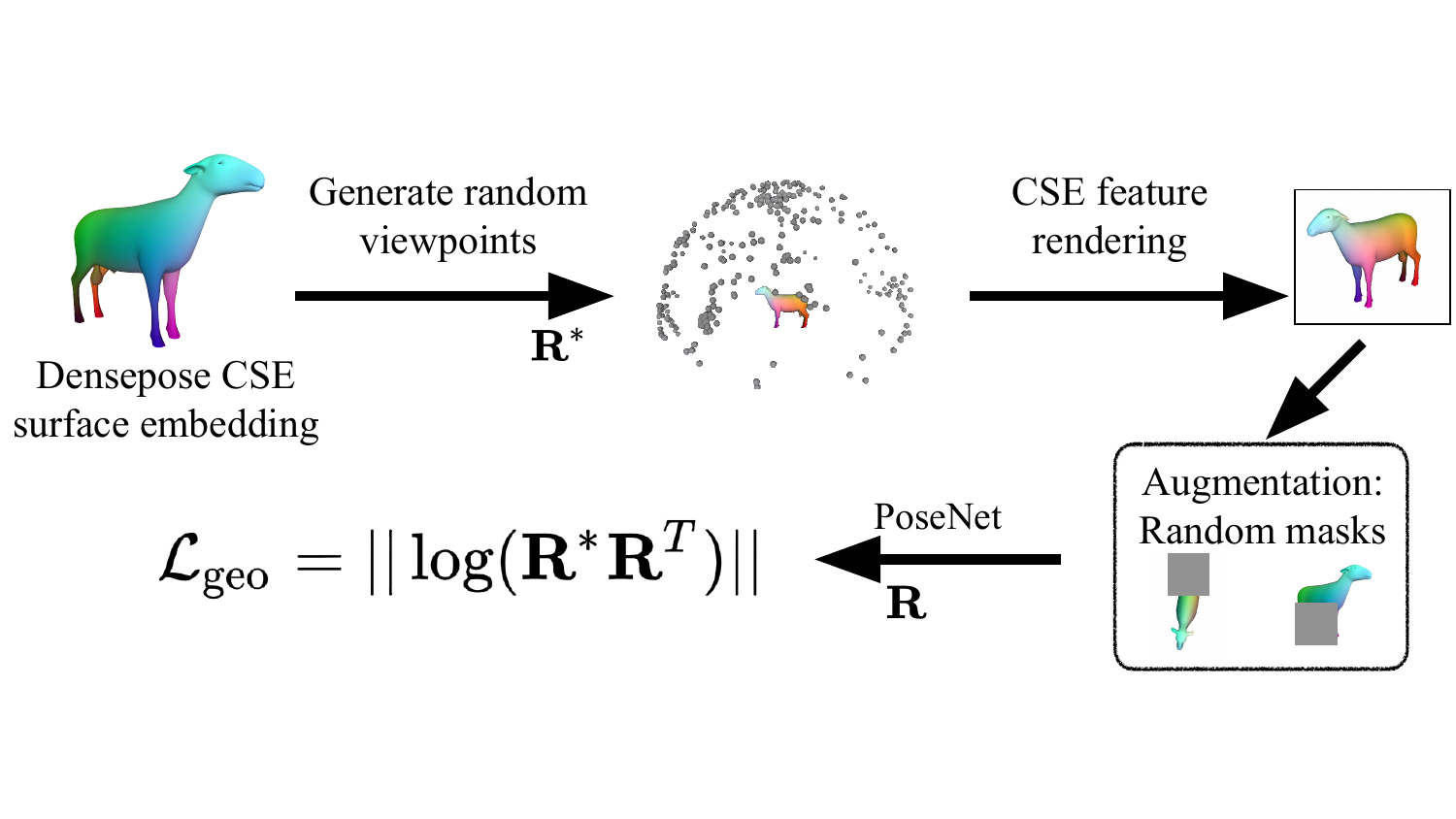}
    \caption{\textbf{Training pipeline of PoseNet.} To train $\mathrm{PoseNet}$, we use DesePose CSE surface embeddings, which is pertained on 2D annotations of human and quadruped animals. We first generate random viewpoints on a sphere that faces the origin. Then we render surface embeddings as 16-channel images. We further augment the feature images with random adversarial masks to improve the robustness to occlusions. Finally, the rotations predicted by $\mathrm{PoseNet}$ are compared against the ground-truth rotations with geodesic distance.}
    \label{fig:posenet-train}
\end{figure}

\noindent{\bf Synthetic dataset genetarion.}
We train separate $\mathrm{PoseNet}$, one for human, and one for quadruped animals. The training pipeline is shown in Fig.~\ref{fig:posenet-train}. Specifically, we render surface features as feature images $\boldsymbol{\psi}_{\textrm{rnd}} \in \mathbb{R}^{112\times 112\times 16}$ given viewpoints ${\bf G}^*=({\bf R}^*,{\bf T}^*)$ randomly generated on a unit sphere facing the origin. We apply occlusion augmentations~\cite{singh2017hide} that randomly mask out a rectangular region in the rendered feature image and replace with mean values of the corresponding feature channels.  The random occlusion augmentation forces the network to be robust to outlier inputs, and empirically helps network to make robust predictions in presence of occlusions and in case of out-of-distribution appearance.

\noindent{\bf Loss and inference.}
We use the geodesic distance between the ground-truth and predicted rotations as a loss to update $\mathrm{PoseNet}$,
\begin{equation}
    \mathcal{L}_{\mathrm{geo}} = ||\log({\bf R}^*{\bf R}^T)||, \quad{\bf R} = \mathrm{PoseNet}(\boldsymbol{\psi}_{\textrm{rnd}}),
\end{equation}
where we find learning to predict rotation is sufficient for initializing the root body pose. In practice, we set the initial object-to-camera translation to be a constant ${\bf T} = (0,0,3)^T$. We run pose CNN on each test video frame to obtain the initial root poses ${\bf G}_0^t=\big({\bf R},{\bf T}\big)$, and compute a delta root pose with the root pose MLP: 
\begin{equation}
   {\bf G}^t= \mathbf{MLP}_{\bf G}({\omega^t_r}) {\bf G}_0^{t}.
\end{equation}

\subsection{\bf Active sampling over $(x,y,t)$}
Inspired by iMAP~\cite{Sucar:etal:ICCV2021}, our ray sampling strategy follows an easy-to-hard curriculum. At the early iterations, we randomly sample a batch of $N^p$ pixels for volume rendering and compute reconstruction losses. At the same time, we optimize a 8-layer MLP function to represent the uncertainty over the image coordinate and frame index: $\hat{\bf U}(x,y,t) = \mathbf{MLP}_{\bf U}(x,y,t)$. The uncertainty MLP is optimized by comparing against the color reconstruction errors in the current forward step:
\begin{equation}
\mathcal{L}_{\bf U} = \sum_{{\bf x},t} \left\|\mathcal{L}_{\textrm{rgb}}({\bf x}^t) - \hat{\bf U}({\bf x}^t)\right\|.
\end{equation}
Note that the gradient from $\mathcal{L}_{\bf U}$ to $\mathcal{L}_{\textrm{rgb}}({\bf x}^t)$ is stopped such that $\mathcal{L}_{\bf U}$ does not generate gradients to parameters besides $\mathbf{MLP}_{\bf U}$. After some optimization steps, we replace half of the samples with \emph{active} samples from pixels with high uncertainties. To do so, we randomly sample $N^{a'}=24576$ pixels, and evaluate their uncertainties by passing their coordinates and frame indices to $\mathbf{MLP}_{\bf U}$. Active samples dramatically improves reconstruction fidelity, as shown in Fig.~\ref{fig:aba-active}. 

\subsection{Optimization details}
\noindent{\bf Canonical 3D grid.}
As mentioned in Sec 3.3, we define a canonical 3D grid ${\bf V}^*\in \mathbb{R}^{20\times20\times 20}$ to compute the matching costs between pixels and canonical space locations. The canonical grid is centered at the origin and axis-aligned with bounds $[x_{\text{min}}, x_{\text{max}}]$, $[y_{\text{min}}, y_{\text{max}}]$, and $[z_{\text{min}}, z_{\text{max}}]$.  The bounds are initialized as loose bounds and are refined during optimization. For every 200 iterations, we update the bounds of the canonical volume as an approximate bound of the object surface. To do so, we run marching cubes on a $64^3$ grid  to extract a surface mesh and then set $L$ as the axis-aligned $(x,y,z)$ bounds of the extracted surface.

\noindent{\bf Near-far planes.}
To generate samples for volume rendering, we dynamically compute the depth of near-far planes $(d^t_n, d^t_f)$ of frame $t$ at each iterations of the optimization. To do so, we compute the projected depth of the canonical surface points $d^t_i = (\Pi^t{\bf G}^t{\bf X}^*_i)_{2}$. The near plane is set as $d_n^t = \min(d_i)-\epsilon_L$ and the far plane is set as $d_f^t=\max(d_i)+\epsilon_L$, where $\epsilon_L = 0.2 \big(\max(d_i) - \min(d_i)\big)$. To avoid the compute overhead, we approximate the surface with an axis-aligned bounding box with $8$ points.

\begin{table}
\caption{Table of hyper-parameters.}
\centering
\begin{tabular}{lll}
\toprule
 Name  & Value & Description \\
 \midrule
 B  & 25 &  Number of bones\\
 $N$  &128    & Sampled points per ray\\
 $N^p$  &6144 & Sampled rays per batch\\
 $(H,W)$ & (512,512) & Resolution of observed images \\
 \bottomrule
\label{tab:hyper-parameters}
\end{tabular}
\end{table}

\noindent{\bf Hyper-parameters.}
We use \href{https://pytorch.org/docs/stable/generated/torch.optim.lr_scheduler.OneCycleLR.html#torch.optim.lr_scheduler.OneCycleLR}{1cycle} learning rate scheduler, which warms-up with a low learning rate to the maximum, and anneals the learning rate to a final learning rate. We apply $lr_{init}=2e-5$, $lr_{max}=5e-4$, $lr_{final}=1e-4$. We refer readers to a complete list of hyper-parameters in Tab.~\ref{tab:hyper-parameters}.

\noindent{\bf Multi-stage optimization}
The final optimization takes three stages, where the optimizable parameters and the loss functions used are different. The first stage uses all the losses and updates all the parameters described in the paper. Typically, the first stage already produces 3D reconstructions with good shape and deformation. The goal of the stage 2 is to improve the articulations (e.g., to correctly articulate the crossing legs for \texttt{cat-pikachiu}) with coordinate gradient descent, where we turn off the reconstruction losses and only use the 2D cycle consistency loss to update the articulation parameters while keeping shape parameters fixed. Finally, stage 3 improves the details of the geometry by active sampling and importance depth sampling while keeping the root body poses fixed.

\noindent{\bf Experiment details}
When running Nerfies on AMA and animated objects, we found using RGB reconstruction loss does not produce meaningful results possibly due to the homogeneous background color. To improve Nerfies results, we provide it with ground-truth object silhouettes, and optimize a carefully balanced RGB+silhouette loss~\cite{yariv2020multiview}.

\section{Additional results}
\subsection{SFM root pose initialization}
COLMAP~\cite{schoenberger2016sfm,schoenberger2016mvs} failed to converge when focused on the deformable object due to violation of rigidity, leading to very few successful registrations (18 over 811 images registered on \texttt{casual-cat}).
A recent end-to-end method, DROID-SLAM~\cite{teed2021droid}, registered all the images but the accuracy is low compared to PoseNet, as shown in Tab.~\ref{tab:quan-pose}.
We also tried SFM to estimate and compensate for the camera motion (using background as rigid anchor), but this did not help to recover the pose of the object due to its global movement w.r.t. to the background.

\begin{table}[h!]
    \vspace{-5pt}
    \caption{\textbf{Evaluation on root pose prediction.} Mean and standard deviation of the rotation error (\textdegree) over all frames ($\downarrow$). We use \ourmethod{}-optimized poses as ground-truth. Rotations are aligned to the ground-truth by a global rotation under \href{https://github.com/scipy/scipy/issues/10862}{chordal L2 distance}.}
    \vspace{-5pt}
    \small
    \centering
    \begin{tabular}{lcccc}
	\toprule
Method & \texttt{c-cat} & \texttt{c-human} & \texttt{ama-human}\\
\midrule
CSE-PoseNet & \boldsymbol{$18.6$}$\pm16.2$ & \boldsymbol{$12.8$}$\pm8.9$ & \boldsymbol{$11.8$}$\pm17.4$\\
DROID-SLAM    & $65.5\pm44.5$ & $55.8\pm39.2$ & $83.6\pm50.5$\\
\bottomrule
\label{tab:quan-pose}
\end{tabular}
\vspace{-20pt}
\end{table}

\subsection{More ablation study}
In Sec.~4.3, we presented qualitative results of diagnostics experiments.
In Tab.~\ref{tab:quan-rebut}, we report the results of other ablations followed by analysis.
\begin{table}[h!]    
    \vspace{-2pt}
    \caption{\textbf{Results on AMA swing and samba.} 3D Chamfer distance (cm, $\downarrow$) and F-score (\%, $\uparrow$) averaged over all frames. }
    \small
    \vspace{-3pt}
    \centering
    \begin{tabular}{lcccc}
	\toprule
Method & \texttt{CD} & \texttt{F@1\%} & \texttt{F@2\%} & \texttt{F@5\%} \\
\midrule
\href{}{number-bone=4} & 9.88 & 28.1 & 52.4 & 84.1\\
\href{}{number-bone=9} & 9.08 & 31.2 & 56.4 & 86.8\\
\href{}{number-bone=16} & {\bf 9.02} & {\bf 31.8} & {\bf 57.2} & 87.2\\
\href{}{number-bone=25} &9.08 &{\bf 31.8} & 57.0 & 87.1\\
\;\;\href{}{--w/o in-surface loss} & 9.14 & 29.9 & 54.8 & 86.7 \\
\;\;\href{}{--quad. embedding} & 9.70 & 29.8 & 54.2 & 85.4\\
\href{}{number-bone=64} &9.18 &31.1 & 56.6 & {\bf 87.5} \\
\href{}{number-bone=100}& 9.11 & 31.4 & 56.7 & 87.3\\
\midrule
\href{}{pose error $\epsilon$=20\textdegree} & {\bf 8.75} & {\bf 30.9} & {\bf 57.0} & {\bf 88.1}\\
\href{}{pose error $\epsilon$=50\textdegree} & 8.91 & 29.8 & 56.1 & {\bf 88.1} \\
\href{}{pose error $\epsilon$=90\textdegree} & 9.91 & 28.4 & 54.8 & 85.7 \\
\midrule
\href{}{coverage=90{\textdegree} (2 vids)} & 10.61 & 29.3& 54.3& 84.1\\
\href{}{coverage=180{\textdegree} (4 vids)} & {\bf 8.94} & {\bf 33.0} & {\bf 59.8} & {\bf 87.9}\\
\href{}{coverage=270{\textdegree} (6 vids)} & 9.09 & 29.8 & 56.1 & 87.6 \\
\midrule
\href{}{active-sample=0\%}&  9.63 & 29.1 & 53.7 & 85.8\\
\href{}{active-sample=25\%}&  {\bf 8.60} & {\bf 32.3} & {\bf 57.9} & {\bf 88.0} \\
\href{}{active-sample=50\%}& 9.14 & 29.9 & 54.8 & 86.7 \\
\bottomrule
\label{tab:quan-rebut}
\end{tabular}
\end{table}

\begin{figure}[ht!]
    \centering
    \vspace{-10pt}
    \includegraphics[width=0.9\linewidth, trim={0 3cm 0 2cm},clip]{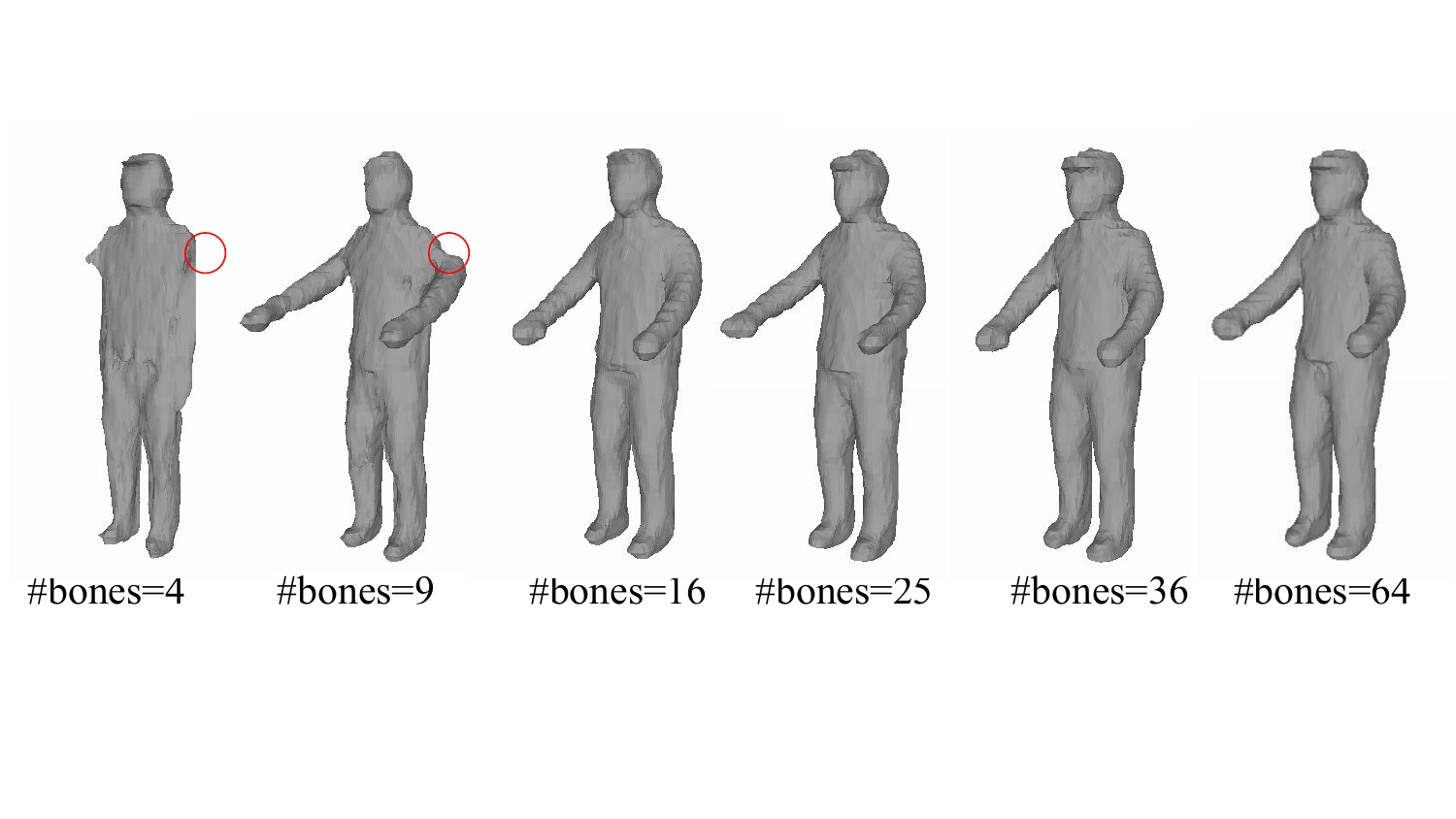}
    \vspace{-10pt}
    \caption{\textbf{Sensitivity to number of bones.}}
    \label{fig:num-bones}
\end{figure}

{\noindent \bf Number and location of bones}
As shown in the first group of Tab.~\ref{tab:quan-rebut} and Fig.~\ref{fig:num-bones}, using too few bones fails to recover all body parts due to over-regularization. Using more than 16 bones produces good reconstructions, but consumes more memory when computing skinning weights. 
Enforcing them to stay close to the surface with a \href{https://www.kernel-operations.io/geomloss/api/pytorch-api.html}{sinkhorn divergence loss} improves the results (Tab.~\ref{tab:quan-rebut}, L16-17).

{\noindent \bf Sensitivity to incorrect initial pose} We inject different levels of Gaussian noise into the initial poses, leading to average rotation errors $\epsilon\in\{20,50,90\}${\textdegree}. As shown in the second group of Tab.~\ref{tab:quan-rebut}, \ourmethod{} is stable up to $50${\textdegree} rotation error.

{\noindent \bf Pre-trained embeddings}
Pre-trained embeddings help BANMo outperform Nerfies, but it is not crucial given good initial root poses ($\epsilon=12.8\pm8.9$\textdegree). As shown in Tab.~\ref{tab:quan-rebut}, using embeddings pre-trained for quadruped animals for human optimization produces slightly worse results. 

{\noindent\bf How much data are needed?}
To reconstruct a complete shape, \ourmethod{} requires all object surface to be visible from at least one frame. Beside completeness, more videos allows to estimate better skinning weights and a more regular motion. We evaluate view coverage in the third group of Tab.~\ref{tab:quan-rebut}.

{\noindent\bf Importance sampling} We use active sampling to avoid sampling from uninformative frames and pixels. 
It consistently improves reconstruction results as shown in the last group of Tab.~\ref{tab:quan-rebut}.

\begin{figure}[ht!]
    \centering
    \includegraphics[width=\linewidth, trim={0cm 1cm 0cm 1cm},clip]{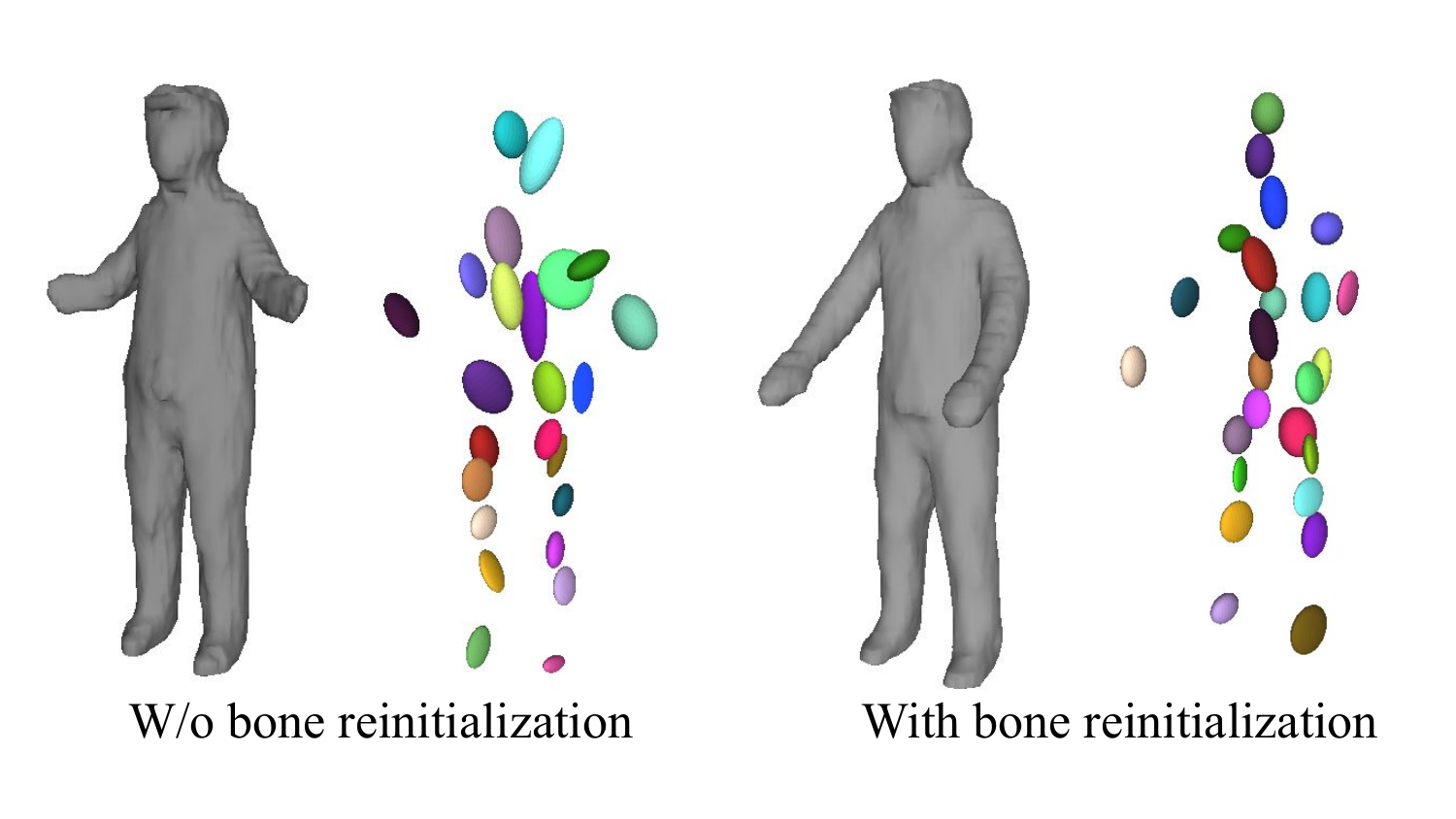}
    \caption{\textbf{Effect of bone re-initialization.} We find it important to re-initialize rest bone parameters after finding a better approximation of object geometry.}
    \label{fig:rebone}
\end{figure}

{\noindent\bf Bone re-initialization} We qualitatively evaluate the effect of rest bone re-initialization, which re-initializes bone parameters according to the current estimation of shape. As shown in Fig.~\ref{fig:rebone}, without re-initializing the bones, the optimization may stuck at bad local optima and the final reconstruction may become less accurate.

\begin{figure}[ht!]
    \centering
    \includegraphics[width=\linewidth, trim={0cm 2cm 0cm 3cm},clip]{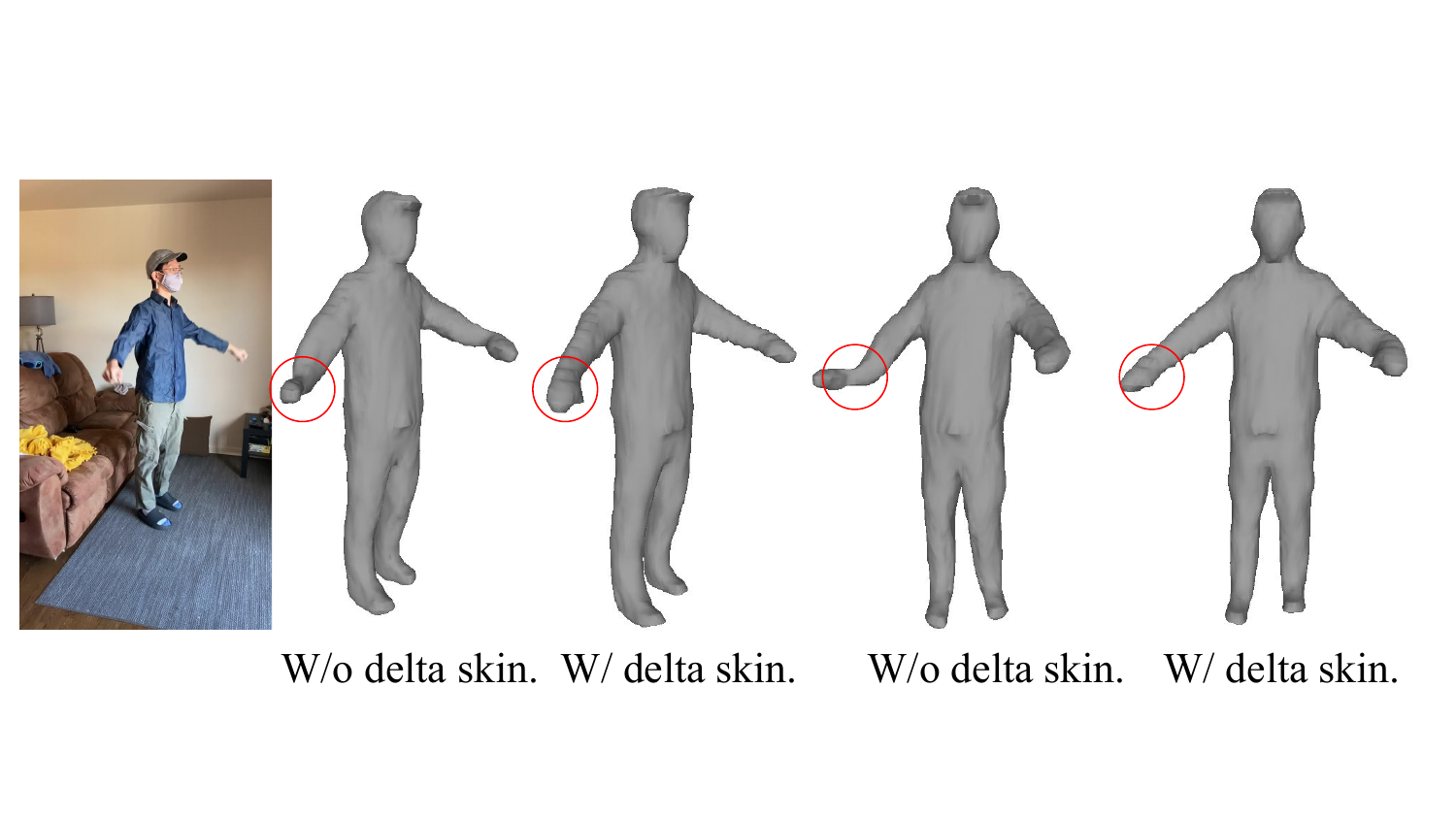}
    \caption{\textbf{Effect of delta skinning weights.} We find it important to learn a point-specific delta skinning weight function to reconstruction motions in high-quality.}
    \label{fig:dskin}
\end{figure}

{\noindent\bf Delta skinning weights} We qualitatively evaluate the effect of delta skinning weights. As shown in Fig.~\ref{fig:dskin}, without learning a delta skinning weights specific to each 3D point, the reconstructed shape and motion may be over-regularized by the 3D Gaussians.

\begin{figure}[ht!]
    \centering
    \includegraphics[width=\linewidth, trim={0cm 3cm 0cm 4cm},clip]{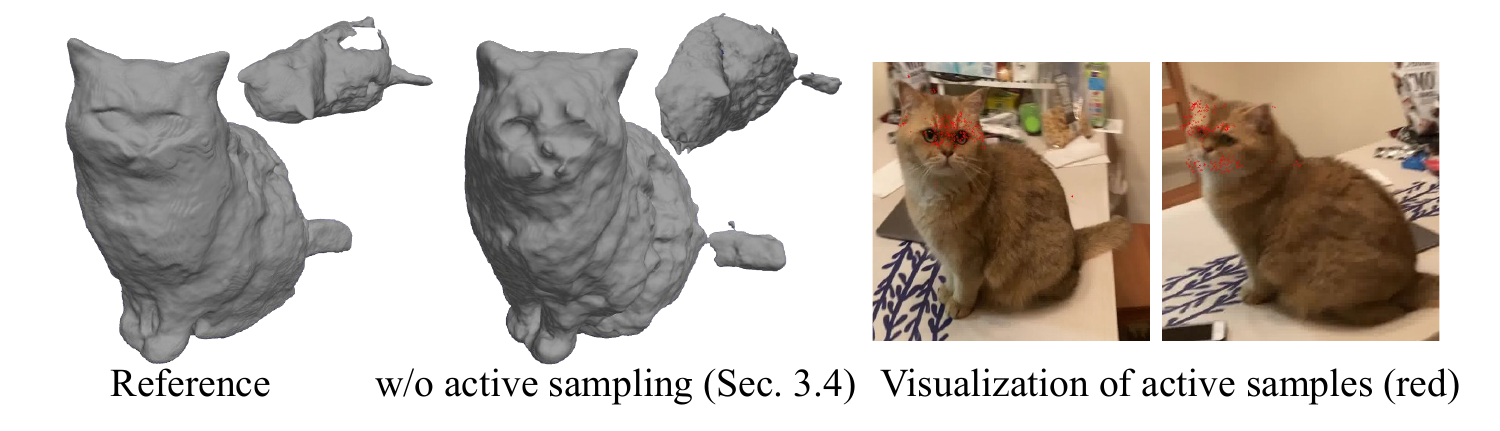}
    \caption{\textbf{Diagnostics of active sampling over $(x,y)$.} With no active sampling, our method converges slower and misses details (such as ears and eyes). %
    Active samples focus on face and boundaries pixels where the color reconstruction errors are higher.}
    \label{fig:aba-active}
\end{figure}

\noindent{\bf Active sampling.}
We show the effect of active sampling on a \texttt{casual-cat} video (Fig.~\ref{fig:aba-active}): removing it results in slower convergence and inaccurate geometry.

\subsection{Qualitative results}
We refer readers to our supplementary webpage for complete qualitative results.

\begin{table*}
\caption{Table of notations.}
\centering
\makebox[\textwidth]{
\begin{tabular}{lll}
\toprule
 Symbol  & Description \\
 \midrule
 \multicolumn{2}{c}{\bf Index}\\
 $t$      & Frame index, $t\in \{1,\dots, T\}$\\
 $b$        &   Bone index $b\in \{1,\dots, B\}$ in neural blend skinning\\
 $i$        &   Point index $b\in \{1,\dots, N\}$ in volume rendering\\

 \midrule
 \multicolumn{2}{c}{\bf Points}\\
 ${\bf x}$& Pixel coordinate {\bf x} = (x,y)\\
 ${\bf X}^t$  & 3D point locations in the frame $t$ camera coordinate\\
 ${\bf X}^*$  & 3D point locations in the canonical coordinate\\
 $\hat{\bf X}^*$  & Matched canonical 3D point locations via canonical embedding\\
 \midrule
 \multicolumn{2}{c}{\bf Property of 3D points}\\
 ${\bf c}\in\mathbb{R}^{3}$& Color of a 3D point\\
 $\sigma\in\mathbb{R}$ & Density of a 3D point\\
 $\boldsymbol{\psi}\in\mathbb{R}^{16}$ & Canonical embedding of a 3D point\\
  ${\bf W}\in\mathbb{R}^{B}$ & Skinning weights of assigning a 3D point to $B$ bones\\
  \midrule
 \multicolumn{2}{c}{\bf Functions on 3D points}\\
  $\mathcal{W}^{t,\leftarrow}({\bf X}^t)$ & Backward warping function from ${\bf X}^t$ to ${\bf X}^*$\\
  $\mathcal{W}^{t,\rightarrow}({\bf X}^*)$ & Forward warping function from ${\bf X}^*$ to ${\bf X}^t$\\
  $\mathcal{S} ({\bf X},\omega_b)$ & Skinning function that computes skinning weights of ${\bf X}$ under body pose $\omega_b$\\
 \midrule
 \multicolumn{2}{c}{\bf Rendered and Observed Images}\\
 ${\bf c} / \hat{\bf c}$      &   Rendered/observed RGB image\\
 ${\bf o} / \hat{\bf s}$      &   Rendered/observed object silhouette image\\
 ${{\mathcal{F}}} / {\hat{\mathcal{F}}}$    &   Rendered/observed optical flow image\\
 \bottomrule
\label{tab:notations}
\end{tabular}
}
\end{table*}

\begin{table*}
\caption{Table of learnable parameters.}
\centering
\makebox[\textwidth]{
\begin{tabular}{lll}
\toprule
 Symbol  & Description \\
 \midrule
 \multicolumn{2}{c}{\bf Canonical Model Parameters}\\
  $\mathbf{MLP}_{\bf c}$           & Color MLP\\
 $\mathbf{MLP}_{\textrm{SDF}}$           & Shape MLP\\
 $\mathbf{MLP}_{\boldsymbol{\psi}}$           & Canonical embedding MLP\\
 \midrule
  \multicolumn{2}{c}{\bf Deformation Model Parameters}\\
 $\boldsymbol{\Lambda}^0\in \mathbb{R}^{3\times3}$                      & Scale of the bones in the zero-configuration (diagonal matrix). \\
 ${\bf V}^0\in \mathbb{R}^{3\times3}$                      & Orientation of the bones in the zero-configuration. \\
 ${\bf C}^0\in \mathbb{R}^3$                      & Center of the bones in the zero-configuration. \\
 $\mathbf{MLP}_{\Delta}$           & Delta skinning weight MLP\\
  $\mathbf{MLP}_{\bf G}$ & Root pose MLP \\
 $\mathbf{MLP}_{\bf J}$  & Body pose MLP\\
 \midrule
 \multicolumn{2}{c}{\bf Learnable Codes}\\
  $\boldsymbol{\omega}^*_b \in \mathbb{R}^{128}$           & Body pose code for the rest pose\\
  $\boldsymbol{\omega}^t_b \in \mathbb{R}^{128}$           & Body pose code for frame $t$\\
  $\boldsymbol{\omega}^t_r\in \mathbb{R}^{128}$           & Root pose code for frame $t$\\
   $\boldsymbol{\omega}^t_{e}\in \mathbb{R}^{64}$         & Environment lighting code for frame $t$, shared across frames of the same video\\
  \midrule
 \multicolumn{2}{c}{\bf Other Learnable Parameters}\\
 $\boldsymbol{\psi}_I$        & CNN pixel embedding initialized from DensePose CSE \\
$\alpha_s$                    & Temperature scalar for canonical feature matching \\
$\beta$                    & Scale parameter that controls the solidness of the object surface\\
   ${\Pi}\in\mathbb{R}^{3\times3}$            & Intrinsic matrix of the pinhole camera model \\
 \bottomrule
\label{tab:parameters}
\end{tabular}
}
\end{table*}